\def\CombinedVersion{}
\documentclass[10pt]{article}

\let\MainAddContentsLine\addcontentsline
\usepackage[preprint]{tmlr}
\let\addcontentsline\MainAddContentsLine

\usepackage{hyperref}
\usepackage{url}
\usepackage{graphicx}
\usepackage{amsmath}
\usepackage{amssymb}
\usepackage{amsthm}
\usepackage{mathtools} 
\usepackage{multicol}
\usepackage{booktabs}

\usepackage[table]{xcolor}
\usepackage{longtable,array}
\usepackage{fvextra}
\usepackage{needspace}
\usepackage[most]{tcolorbox}

\title{Large Language Models Extend Known Research But Do Not Create New Research Directions}

\author{\name Zhihua Liu\thanks{Equal contribution.} \email zliu7@ed.ac.uk \\
      \addr Causality in Healthcare AI Hub (CHAI) \\
      School of Engineering, University of Edinburgh
      \AND
      \name Kurt Butler\footnotemark[1] \email kbutler2@ed.ac.uk \\
      \addr Causality in Healthcare AI Hub (CHAI) \\
      School of Engineering, University of Edinburgh
      \AND
      \name Damian Machlanski\footnotemark[1] \email d.machlanski@ed.ac.uk \\
      \addr Causality in Healthcare AI Hub (CHAI) \\
      School of Engineering, University of Edinburgh
      \AND
      \name Danilo Bueno \email danilohrbueno@gmail.com \\
      \addr Causality in Healthcare AI Hub (CHAI) \\
      School of Engineering, University of Edinburgh
      \AND
      \name Sotirios A. Tsaftaris \email S.Tsaftaris@ed.ac.uk \\
      \addr Causality in Healthcare AI Hub (CHAI) \\
      School of Engineering, University of Edinburgh}

\DeclarePairedDelimiter{\norm}{\lVert}{\rVert}

\begin{document}

\maketitle

\begin{abstract}
Autonomous research systems increasingly use large language models (LLMs) to generate scientific hypotheses, which human reviewers often rate as highly novel~\citep{si2025can,si2026ideation, jiang2026hindsight}.
Yet novelty can mean either extending an established line of inquiry or opening a new one. Whether LLMs can do the latter remains unclear~\citep{bao2026contemporary}.
We distinguish these possibilities by measuring hypotheses along two dimensions: distance from existing work and direction beyond it.
We embed 26.9 million human-authored papers across 1,325 topics in six natural-science fields and 530,000 hypotheses generated by eight LLMs in a shared scientific embedding space~\citep{singh2023scirepeval}. We quantify distance using established metrics~\citep{tang2026ai} and introduce the orthogonal novelty score (ONS), which measures how much a hypothesis points outside directions already spanned by the literature. 
LLM-generated hypotheses lie farther from their nearest existing paper than human-authored comparator papers (up to 16\%), yet their median ONS is 9--16\% lower. 
Across models, fields, context configurations, feedback loops and personas, the overall pattern is robust: LLMs extend established research trajectories but do not open new directions as reliably as human-authored papers. 
These findings show why distance-based novelty alone can overstate the originality of scientific hypotheses generated by LLMs and provide a framework for evaluating directional novelty separately.\footnote{Project page:~\url{https://chai-uk.github.io/arcadia-project-page/}}
\end{abstract}

\newpage
LLM-based autonomous research systems are becoming commonplace. In the past, the use of LLMs in science was limited to specific tasks~\citep{singhal2023large, guo2023can}. Now, end-to-end platforms guided by LLM-generated hypotheses are increasingly being deployed~\citep{tang2026ai, baek2025researchagent, li2025chain, gottweis2026accelerating, ghareeb2026multi}. AI-designed SARS-CoV-2 
nanobodies have recently been reported, with experimental 
validation confirming biological activity~\citep{swanson2025virtual}. The question of whether these systems can drive genuine scientific 
discovery is therefore no longer hypothetical. In early evaluations, reviewers rate LLM-generated hypotheses as highly novel~\citep{si2025can, abdel2025scientific}. Recent benchmark work further shows that scientific idea generation is not well predicted by general reasoning performance, motivating dedicated measures of scientific creativity and originality~\citep{ruan2026evaluating}. Yet whether these ideas extend established research directions or introduce new ones remains unclear.

This uncertainty is reinforced by other studies that suggest LLMs make science narrower. \cite{hao2026artificial} report that AI tool use expands scientists’ impact but contracts their focus.~\cite{tang2026ai} find that LLM-generated ideas are more conceptually concentrated than human-authored papers.~\cite{luyten2026reasoning} show theoretically that novel hypothesis generation is penalized by the need to optimize for correctness, which reduces the diversity of acceptable hypotheses. These findings create an apparent contradiction: do LLM-based research agents broaden the horizon of science, or do they constrain it?

We argue that this paradox arises from a failure to distinguish two dimensions of novelty. In the study of creativity, \cite{boden2004creative} distinguishes exploration within an existing conceptual space from transformational creativity that changes the space itself.~\cite{jaksland2021teaching} connects this distinction to Kuhn’s contrast between normal and revolutionary science~\citep{kuhn1970structure}. Our distinction between distance and direction is intended as an embedding-space analog of this difference. A scientific hypothesis can be distant from existing work while remaining entirely within the established conceptual directions. Normal science extends existing research directions, occasionally to a considerable distance, whereas conceptual change requires new directions. Empirical studies of scientific search identify a persistent tension between exploitation of established knowledge and riskier innovative strategies~\citep{foster2015tradition}. Existing measures of novelty do not distinguish between these two notions, leaving open whether the narrowing reported in these studies represents a superficial tuning problem or a fundamental structural constraint.

Answering this requires measuring direction separately from distance. Bibliometric studies have used semantic embeddings to quantify scientific novelty, including the semantic distance among prior works combined by a paper~\citep{shibayama2021measuring}. Embedding-based novelty assessments commonly evaluate a paper using either its Euclidean distance from existing work or its cosine similarity to individual published papers~\citep{si2025can,tang2026ai,hao2026artificial}. Both metrics conflate \textit{how far} with \textit{where} a hypothesis is placed. A hypothesis that extrapolates along an established axis of a scientific domain can lie far away from all published papers while introducing no new conceptual directions. Pairwise cosine similarity has a further weakness. A hypothesis that recombines several existing papers may be dissimilar to each of them individually, yet be almost fully explained by them jointly, highlighting that per-paper comparisons may overstate a paper's novelty. We therefore introduce the orthogonal novelty score (ONS), which measures not how far a hypothesis lies from existing work, but how much of it points in directions the existing literature has not yet explored. A hypothesis whose embedding is a combination of the reference embeddings scores zero, however distant it may be, as high ONS requires content outside the directions spanned by the literature.

\begin{figure}[t]
    \centering
    \includegraphics[width=\linewidth]{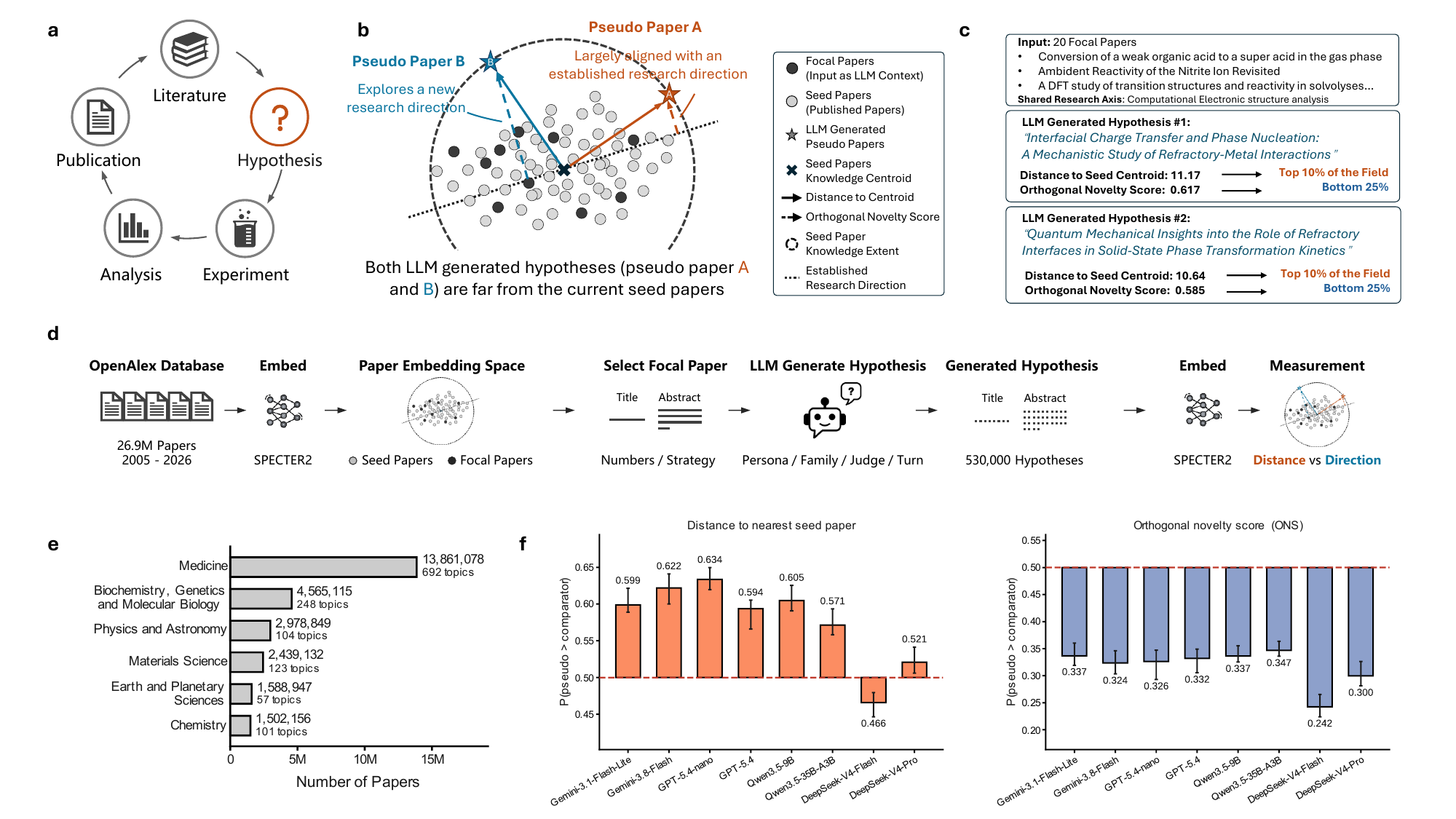}
    \caption{\textbf{LLMs place hypotheses farther from the published literature than comparator papers do, yet along fewer new research directions.}~\textbf{a}, Illustration of hypothesis generation within an automated research loop.~\textbf{b}, Distance versus direction novelty of an LLM-generated hypothesis.~\textbf{c}, A running example of an LLM-generated hypothesis that falls within the top decile for distance extension and the bottom quartile for direction orthogonality. ~\textbf{d}, Overview of our generation and evaluation pipeline.~\textbf{e}, Scale of the comparison corpus.~\textbf{f}, Comparison between generated hypotheses and human-authored papers across three evaluation metrics.}
    \label{fig:1}
\end{figure}

To test distance and direction separately, we took 26.9 million human-authored papers from OpenAlex~\citep{priem2022openalex,openalex} published between 2005 and 2026 spanning 1,325 research topics in six natural-science fields and embedded them into a shared 768-dimensional embedding space~\citep{singh2023scirepeval} (Fig.~\ref{fig:1}). We refer to these embedded human-authored papers as \textbf{seed papers}. For each topic, an LLM ideator received a subset of these papers as context (\textbf{focal papers}) and generated hypotheses in the form of titles and abstracts (\textbf{pseudo papers}), which we compared with approximately 1,000 held-out human-authored papers from the same topic (\textbf{comparator papers}). We evaluated generated and comparator papers using three measures: distance to the centroid of the topic's seed papers, distance to the nearest seed paper in the topic, and ONS, which measures the proportion of a paper’s centered embedding that lies outside the subspace spanned by the existing literature within a topic.

Across eight state-of-the-art LLMs and 530,000 generated hypotheses, distance and direction behaved differently: LLM hypotheses exceed human-authored papers on distance-based novelty but fall short on direction-based novelty. This pattern holds across model families and most tested interventions. Selecting context papers from the far edge of a topic brought directional novelty to parity with human-authored papers for the two tested models, but only while pushing generated hypotheses substantially farther from existing work. No tested configuration produced significantly greater directional novelty than human-authored papers. These results identify a specific and reproducible limitation on the contribution of LLMs to scientific discovery.

\section*{Results}
We compare the pseudo papers of each topic with the comparator papers of the same topic, scored on the same three metrics. For each metric, we report the win rate $P(\text{pseudo} > \text{comparator})$, the probability that a randomly drawn pseudo paper scores higher than a comparator paper drawn at random from the same topic. A win rate $P(\text{pseudo} > \text{comparator}) < 0.5$ indicates that pseudo papers tend to score lower than comparator papers on that metric. We also report the ratio of the median pseudo paper to the median comparator paper, which expresses a difference as a percentage. Both quantities are computed within each topic. We summarize them as the median of the six per-field medians, so each field counts once regardless of how many topics it contains. For the field-scale analysis, each field's median is itself the median of its per-subfield medians. Confidence intervals are 95\% bootstrap intervals that resample topics within each field (within each subfield for the field-scale analysis). See \textit{Methods} for more details.

We first present the results from a large-scale comparison across eight state-of-the-art LLMs and six natural-science research fields (Fig.~\ref{fig:2}). We then test two categories of interventions that most systems using LLMs for automated research can control, including the context the LLM reads, through the number and the selection strategy of focal papers (Figs~\ref{fig:3} and~\ref{fig:4}), and the process that refines LLM-generated hypotheses, through a multi-turn dialogue with an LLM judge and the personas given to the two roles (Figs~\ref{fig:5} and~\ref{fig:6}). Further robustness checks and ablation studies are provided in the Supplementary Information.



\subsection*{LLM Generated Hypotheses Open Fewer New Research Directions}

By separating distance from direction, we ask: \textit{Do LLM-generated hypotheses fall short of published work, and does the answer depend on the field or on the model?} We conduct a large-scale comparison across eight LLMs from four model families (Gemini, GPT, Qwen and DeepSeek). Each LLM generates 50 pseudo papers per research topic given 20 focal papers of that topic as context, yielding a total of 530,000 LLM-generated hypotheses across 1,325 research topics.

\begin{figure}[!ht]
    \centering
    \includegraphics[width=0.91\linewidth]{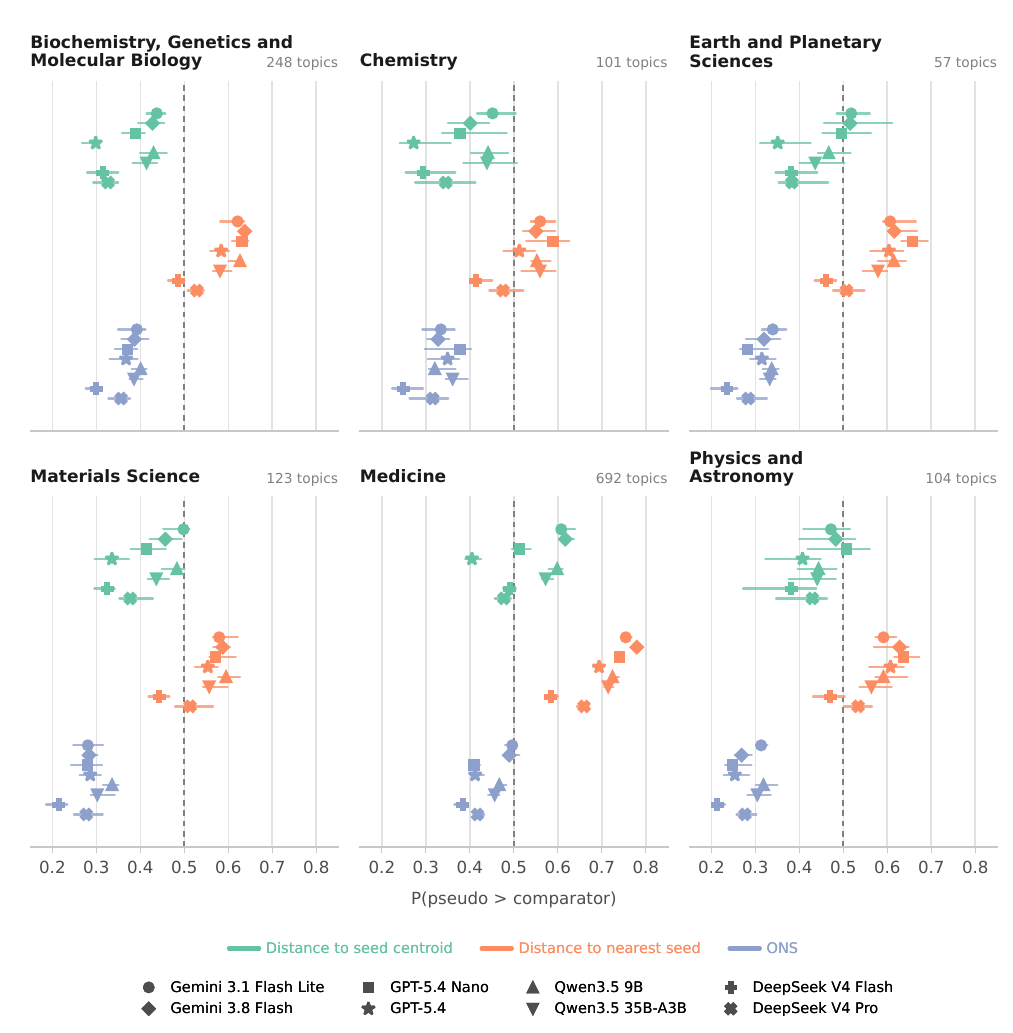}
    \caption{\textbf{LLM-generated hypotheses open fewer new research directions than comparator papers across fields and models.} Pseudo papers generated by eight LLMs (marker shapes), each using 20 focal papers per topic, are compared with held-out comparator papers from the same topic across six fields. The three metrics are distance to the topic centroid (green), distance to the nearest seed paper (orange) and orthogonal novelty score (ONS; blue). The x-axis shows the probability that a pseudo paper exceeds a comparator paper on each metric, with 0.5 indicating parity. Markers show medians across topics within each field, with 50 hypotheses generated per topic and model. Bars show 95\% confidence intervals.}
    \label{fig:2}
\end{figure}

Most LLMs generate hypotheses farther from their nearest seed paper than comparator papers are from theirs (win rate above 50\% in 42 of 48 field–model combinations, $P < 0.05$). By contrast, distance to the topic centroid is mostly shorter for pseudo papers than for comparator papers, with median win rates of 34--48\% across fields ($P \leq 0.032$ in seven of the eight models; $P = 0.14$ for Gemini-3.1-Flash-Lite). Most LLMs therefore move generated hypotheses away from their nearest neighbors without moving them farther from the center of the topic. The hypotheses fill gaps within the topic rather than push its boundary outward (Fig.~\ref{fig:2}). The shortfall is larger for direction than for distance across all LLMs (median win rate 24--35\% on ONS against 34--48\% on centroid distance, $P<0.001$). The median pseudo paper has an ONS 9--16\% lower than the median comparator paper ($P<0.001$). This contraction appeared in all six fields and all eight LLMs: the ONS win rate was below 50\% in all 48 field–model combinations ($P < 0.001$). When a comparator paper and a pseudo paper are drawn at random from the same topic, the comparator paper has the higher ONS in 65--76\%.



Moving to the larger model of a family did not remove the ONS contraction in any family. Gemini-3.8-Flash had a win rate of 32.4\%, compared with 33.7\% for Gemini-3.1-Flash-Lite. GPT-5.4 achieved a win rate of 33.2\%, which is statistically unchanged compared with 32.6\% for GPT-5.4-Nano. The same held for Qwen3.5-35B-A3B (win rate 34.7\%) versus Qwen3.5-9B (33.7\%). DeepSeek-V4-Pro improved on DeepSeek-V4-Flash (30\% versus 24.2\%), but remained well below comparator papers. These comparisons between released models within a family differ in more ways than their number of parameters.

These findings show that the contraction in research directions is shared by every model and field we examined, whereas how far the hypotheses travel depends on both. Since the contraction persists regardless of model, we next asked whether it depends on \textit{how} models are used. 


\subsection*{Context Moves Hypotheses Outward More Than Into New Directions}
\label{subsec:context}

The first type of intervention we examined was model input (also commonly referred to as LLM \textit{context}), which in our setting is the set of focal papers. As the large-scale analysis in Fig.~\ref{fig:2} showed little sensitivity to model choice, we focused here on two representative models to reduce computational demand: Gemini-3.1-Flash-Lite and GPT-5.4-Nano. We varied the context in two ways: the number of focal papers and how they are selected. All interventions were performed across 86 topics, taking from each of the 86 subfields the topic with the largest number of seed papers, and generating 50 pseudo papers per topic in every setting.

\begin{figure}[t]
    \centering
    \includegraphics[width=0.95\linewidth]{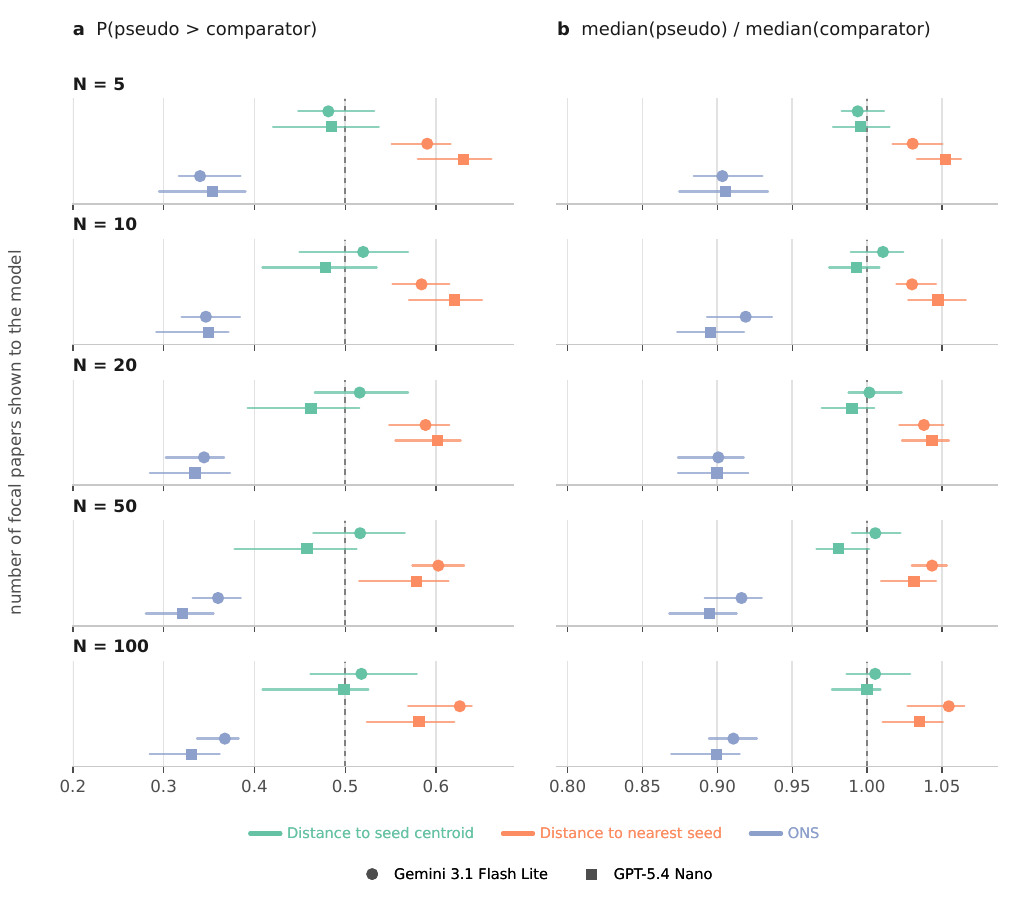}
    \caption{\textbf{The number of focal papers does not change how generated hypotheses compare with comparator papers.} Pseudo papers generated from $N \in \{5, 10, 20, 50, 100\}$ focal papers (rows) are compared with held-out comparator papers of the same topic on three metrics: distance to the topic centroid (green), distance to the nearest seed paper (orange) and ONS (blue). \textbf{a}, Probability that a pseudo paper exceeds a comparator paper on each metric. \textbf{b}, Ratio of the median pseudo paper to the median comparator paper. Markers show the median of six field medians over 86 topics, with 50 hypotheses generated per topic and setting and up to 1,000 held-out comparator papers per topic. Bars show 95\% confidence intervals.}
    \label{fig:3}
\end{figure}

\textbf{Number of focal papers.} We first varied the number of focal papers provided to the LLM from 5 to 100 (Fig.~\ref{fig:3}, $N \in \{5, 10, 20, 50, 100\}$). None of the three metrics changed with the number of focal papers. Generated hypotheses were 3--5\% farther from their nearest seed paper than comparator papers were at every $N$ for both models (Fig.~\ref{fig:3}a,b; win rate 58--63\%; $P \leq 0.015$ at every $N$), while their distance to the topic centroid stayed within 2\% of that of comparator papers and did not differ significantly at any $N$ (win rate 46--52\%; $P \geq 0.13$). The median pseudo paper had an ONS 8--11\% lower than the median comparator paper at every $N$ for both models (win rate 32--37\%; $P < 0.001$ at every $N$), and this contraction stayed consistent across the range of numbers of focal papers.

\begin{figure}[t]
    \centering
    \includegraphics[width=0.95\linewidth]{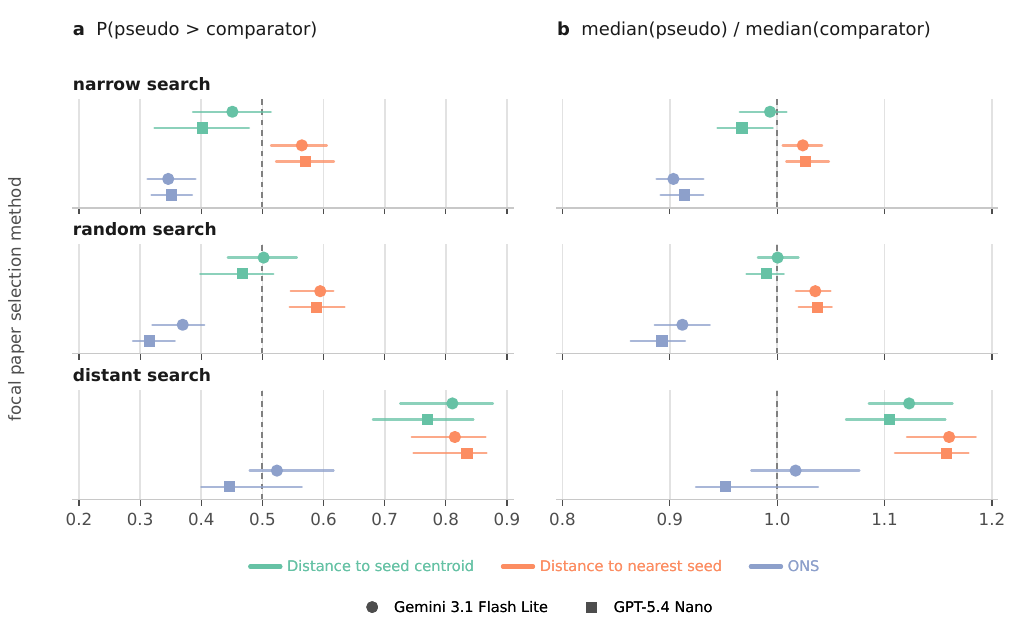}
    \caption{\textbf{Selecting focal papers far from the topic centroid displaces generated hypotheses and leaves no detected ONS difference from comparator papers.} Pseudo papers generated from 20 focal papers selected by three strategies (rows): the 20 papers closest to the topic centroid (\textit{narrow search}), 20 papers drawn at random (\textit{random search}) and the 20 papers farthest from the centroid (\textit{distant search}). They are compared with held-out comparator papers of the same topic on distance to the topic centroid (green), distance to the nearest seed paper (orange) and ONS (blue). \textbf{a}, Probability that a pseudo paper exceeds a comparator paper on each metric. \textbf{b}, Ratio of the median pseudo paper to the median comparator paper. Markers show the median of six field medians over 86 topics, with 50 hypotheses per topic per strategy compared against up to 1,000 held-out comparator papers per topic, the same for every strategy. Bars show 95\% confidence intervals.}
    \label{fig:4}
\end{figure}

\textbf{Selection strategy of focal papers.} We next asked whether \textit{which} papers are selected matters. With $N=20$ fixed, we selected focal papers by three strategies based on Euclidean distance in the embedding space: the 20 papers closest to the topic centroid (narrow search), 20 papers drawn at random (random search), and the 20 papers farthest from the centroid (distant search). The results are summarized in Fig.~\ref{fig:4}. With the farthest papers as context, LLM-generated hypotheses were located 11--12\% farther from the topic centroid and 16\% farther from their nearest seed paper than comparator papers (Fig.~\ref{fig:4}a,b, win rate 77--84\%; $P < 0.001$  for both distances and both models). With random selection, their distance to the centroid did not differ significantly from that of comparator papers for either model. With the closest papers as input, it did not differ for Gemini-3.1-Flash-Lite and was 3\% shorter for GPT-5.4-Nano (win rate 40\%). Distance to the nearest seed paper was greater than that of comparator papers under all three strategies for both models. The choice of focal papers also affected the direction measurement. With the closest or randomly selected papers as input, ONS was 9--11\% lower than that of comparator papers for both models (win rate 32--37\%, $P < 0.001$), matching the contraction observed at every $N$ in Fig.~\ref{fig:3}. Only with the farthest papers did ONS no longer differ significantly from that of comparator papers for either model: it was 2\% higher for Gemini-3.1-Flash-Lite (win rate 52\%) and 5\% lower for GPT-5.4-Nano (win rate 45\%).

Collectively, changing the number of focal papers had no effect on the contraction, and changing their selection strategy affected how far pseudo papers were displaced. Only the farthest papers brought ONS to the level of that of comparator papers, and only by placing hypotheses 11--16\% farther from the topic centroid and the nearest seed paper. Neither intervention made pseudo papers outperform comparator papers. We therefore turned to the generation process itself.

\subsection*{LLM Feedback and Personas Change Hypotheses Without Opening New Directions}

Autonomous research systems commonly pair an LLM ideator that generates research hypotheses with an LLM judge that reviews them, and feed the reviews back to the ideator for the next round of generation~\citep{lu2026towards}. We reproduced this design as a five-turn dialogue between an LLM ideator and an LLM judge, using the same two LLMs and the same 86 topics as in Figs.~\ref{fig:3} and \ref{fig:4}. In every dialogue turn, the ideator received the same 50 focal papers and generated 20 pseudo papers per topic. From the second turn onward, the judge first reviewed the previous batch of hypotheses as a whole, naming its weaknesses, its overrepresented and missing themes, and concrete suggestions for the next batch. The ideator then generated a new batch conditioned on the original set of focal papers, the feedback from the judge, and the titles of all earlier pseudo papers, which the ideator was told not to repeat in future rounds. We ran two variants of the dialogue, swapping which model acts as ideator and which as judge. In addition to the LLM feedback intervention as described above, we also changed the behavior of LLMs via persona prompts~\citep{kong2024better}, a mechanism routinely used to assign roles in automated research systems~\citep{lu2026towards}.

\begin{figure}[t]
    \centering
    \includegraphics[width=0.95\linewidth]{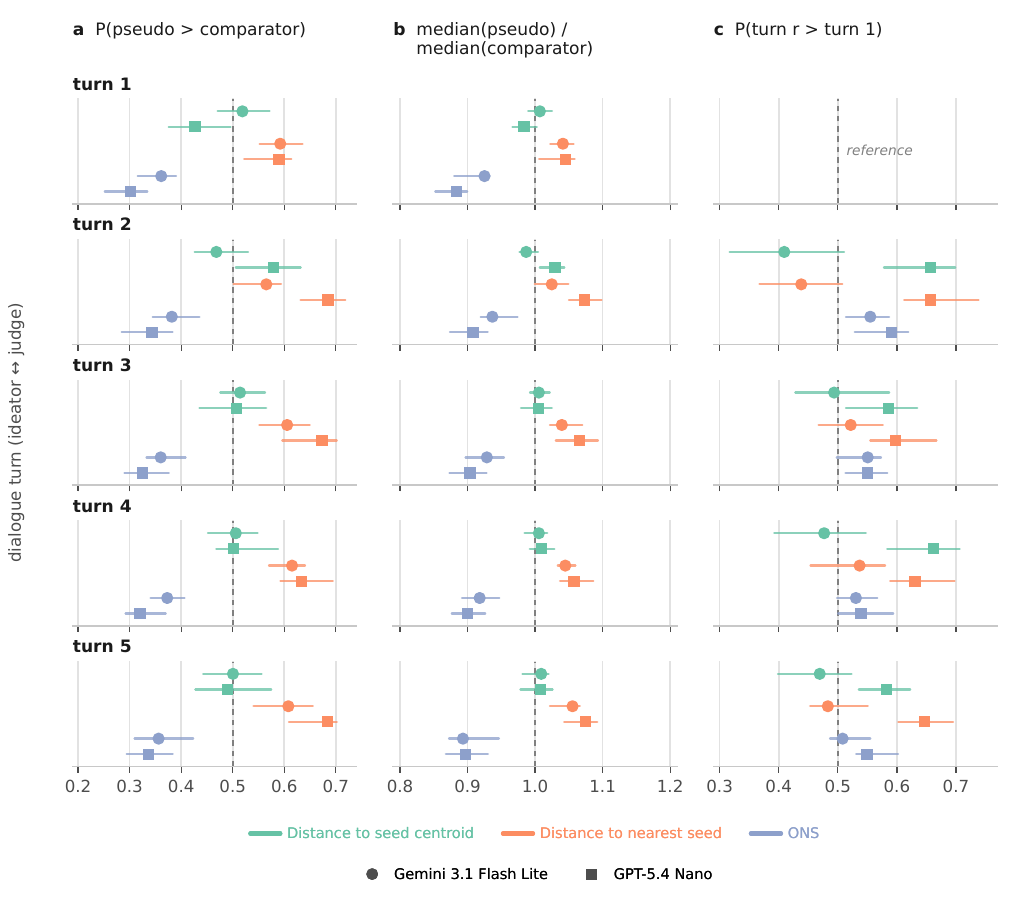}
    \caption{\textbf{LLM feedback changes generated hypotheses without opening new directions.} Pseudo papers generated over five ideator--judge dialogue turns (rows), with both agents given neutral persona prompts. The judge first gives feedback after turn 1. Pseudo papers are compared on distance to the topic centroid (green), distance to the nearest seed paper (orange) and ONS (blue). \textbf{a}, Probability that a pseudo paper exceeds a held-out comparator paper on each metric (win rate). \textbf{b}, Ratio of the median pseudo paper to the median comparator paper. \textbf{c}, Probability that a pseudo paper from turn $r$ exceeds one from turn 1 of the same topic (turn 1 is the reference). Markers show the median of six field medians over 86 topics, with 50 focal papers and 20 hypotheses per topic per turn, compared against up to 1,000 held-out comparator papers per topic, the same at every turn. Bars show 95\% confidence intervals.}
    \label{fig:5}
\end{figure}

\textbf{LLM feedback.} With neutral persona prompts for both the ideator and the judge, we compared the pseudo papers generated in each turn with comparator papers (Fig.~\ref{fig:5}a,b) and with the ideator's own first turn (Fig.~\ref{fig:5}c). Against comparator papers, the median pseudo paper had an ONS 6--12\% lower than the median comparator paper (Fig.~\ref{fig:5}a,b; win rate 30--38\%; $P < 0.001$ at each of five turns for both models), while pseudo papers were 2--7\% farther from their nearest seed paper than comparator papers were (win rate 57--69\%; $P \leq 0.037$ at every turn for both models). Relative to turn one of each model, feedback moved the later hypotheses of GPT-5.4-Nano farther from the topic centroid and from their nearest seed paper (Fig.~\ref{fig:5}c; win rate 58--66\%, $P \leq 0.019$ at each of turns 2 to 5). Those of Gemini-3.1-Flash-Lite did not change significantly on either distance (win rate over turn 1 41--54\%, $P \geq 0.09$ at each of turns 2 to 5).
The change over the first turn on ONS was small. For GPT-5.4-Nano, ONS was higher than at turn 1 at turns 2, 3 and 5 (win rate over turn 1 55–59\%, $P \leq 0.017$) but not significantly at turn 4 (54\%, $P = 0.052$). For Gemini-3.1-Flash-Lite, it was higher only at turn 2 (56\%, $P = 0.010$), not at turns 3–5 (51–55\%, $P \geq 0.06$). These small gains did not close the gap as ONS stayed 6–12\% below that of comparator papers at every turn, including hypotheses generated from the last feedback turn.

\begin{figure}[t]
    \centering
    \includegraphics[width=0.9\linewidth]{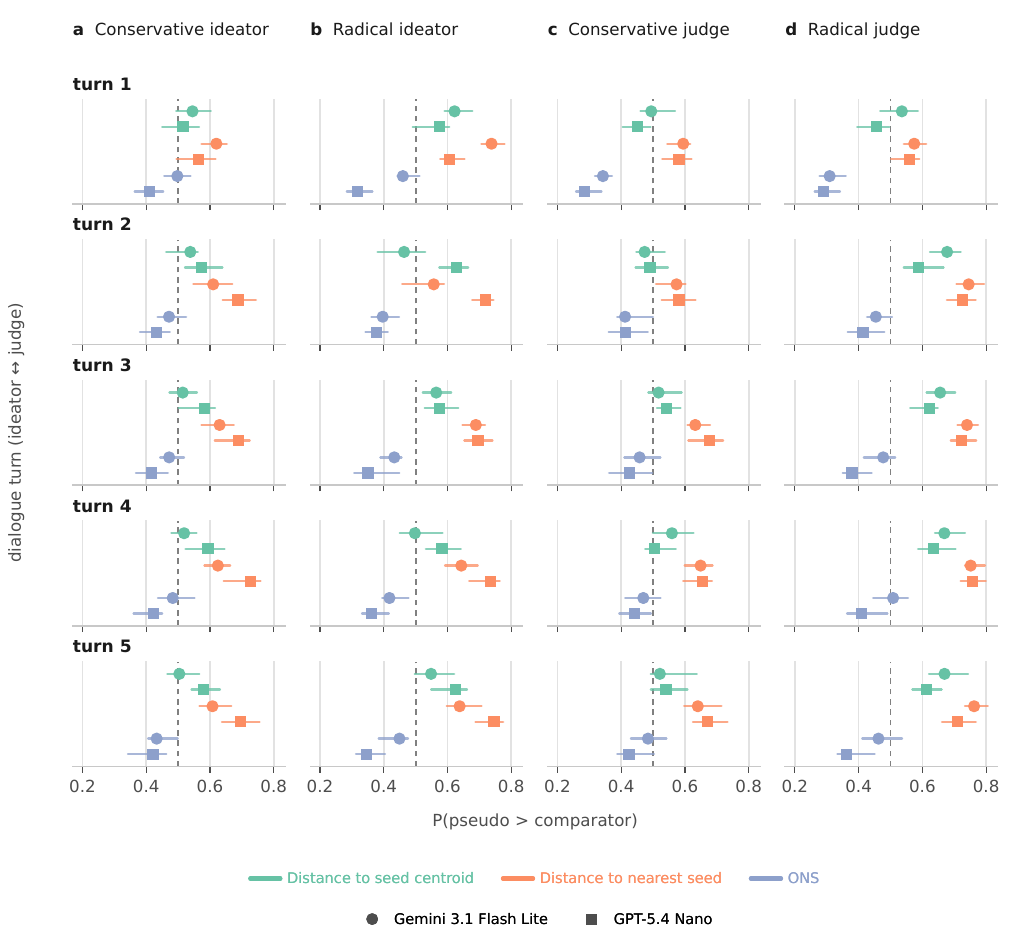}
    \caption{\textbf{LLM personas alter generated hypotheses without raising directional novelty above comparator papers.} Pseudo papers are generated over five ideator--judge dialogue turns under four persona configurations: \textbf{a}, conservative ideator (neutral judge); \textbf{b}, radical ideator (neutral judge); \textbf{c}, conservative judge (neutral ideator); \textbf{d}, radical judge (neutral ideator). Each panel shows, over five ideator--judge dialogue turns (rows), the probability that a pseudo paper exceeds a held-out comparator paper on distance to the topic centroid (green), distance to the nearest seed paper (orange) and ONS (blue). In c and d the ideator is the same, so the two panels differ at turn 1 only by sampling. Markers show the median of six field medians over 86 topics, with 50 focal papers and 20 hypotheses per topic per turn, compared against up to 1,000 held-out comparator papers per topic, shared by all four panels. Bars show 95\% confidence intervals.}
    \label{fig:6}
\end{figure}

\textbf{Personas.} The LLM feedback cycle in Fig.~\ref{fig:5} used neutral persona prompts, which request hypotheses that bridge the topic to neighboring areas or adjacent disciplines. To test the effect of personas, we repeated the five-turn feedback with a non-neutral persona given to one role at a time, either the ideator or the judge, while the other role kept its neutral prompt. We experimented with two types of personas: \textit{conservative} and \textit{radical}. The conservative persona asks for sharper and more rigorous hypotheses that extend the focal papers. The radical persona asks for the opposite, using terms close to what ONS measures: a hypothesis should bridge the topic with a distant discipline, invert an assumption the focal papers take for granted, attack a problem they treat as solved, or propose a method or framing the field has not yet seen. This yields four configurations depending on the role in the dialogue and selected persona: \textit{conservative ideator}, \textit{radical ideator}, \textit{conservative judge}, and \textit{radical judge} (Fig.~\ref{fig:6}a--d).

No persona raised the ONS of generated hypotheses above that of comparator papers. Across all 40 configurations, the median pseudo paper had an ONS between 12\% lower and 1\% higher than the median comparator paper (Fig.~\ref{fig:6}; win rate 28--51\%), and ONS was significantly lower than that of comparator papers in most of them and significantly higher in none ($P < 0.05$ in 27 of 40 configurations). Meanwhile, the distance to the nearest seed paper was greater than that of comparator papers in most configurations (win rate 56--76\%; $P < 0.05$ in 38 of 40 configurations). Explicitly asking for new directions using a \textit{radical} persona did not raise ONS. Compared directly at the same feedback turn, the pseudo papers generated by the radical ideator had a lower ONS than those of the conservative ideator at every turn for both models (win rate of radical over conservative 39--48\%), significantly so from the second turn onward for Gemini-3.1-Flash-Lite ($P \le 0.035$) and at the first turn for GPT-5.4-Nano ($P < 0.001$).

The persona of the judge had the largest effect on distance. From the second turn, when the judge first provides feedback, the win rate in terms of distance to the nearest seed paper was 71--76\% under the radical judge ($P < 0.001$) against 57--68\% under the conservative judge (Fig.~\ref{fig:6}c,d; $P \leq 0.023$). Compared directly, the radical judge placed generated hypotheses farther from their nearest seed paper than the conservative judge at every turn from the second onward for both models (win rate 61--77\%; $P \leq 0.012$), whereas the two ideator personas differed far less and not always in the same direction (win rate 37--60\%). The judge's persona, however, made little difference to ONS. From the second turn onward, ONS under the radical judge did not differ significantly from that under the conservative judge (win rate 45--55\%; $P \geq 0.23$), except for GPT-5.4-Nano at the final turn, where it was lower (win rate 45\%; $P = 0.048$). Collectively, LLM feedback and personas affected the location of pseudo papers in the embedding space, most strongly through the judge, but none raised their ONS above that of comparator papers.

\section*{Discussion}
Here we perform a large-scale empirical measurement of the novelty of LLM-generated scientific hypotheses, separating two properties that novelty metrics commonly conflate: how far a hypothesis lies from published work, and whether it opens a new research direction. Evaluating 530,000 hypotheses generated by eight large language models against 26.9 million papers across 1,325 topics in six natural-science fields reveals that these two dimensions are distinct. LLM-generated hypotheses lie farther from their nearest seed paper than published (comparator) papers do. However, their median ONS remains 9--16\% lower than that of published (comparator) papers. Increasing context size, pairing the ideator with an LLM judge, changing personas and using larger models did not remove this directional contraction. Selecting context papers from the topic boundary brought ONS to the level of that of comparator papers, but only by displacing hypotheses farther from published work, and no intervention raised ONS significantly above that of comparator papers. These findings reconcile conflicting claims in recent literature, showing that reports of high novelty in language-model ideas~\citep{si2025can} and evidence of narrowed scientific exploration~\citep{hao2026artificial,tang2026ai} are not mutually exclusive.

These empirical insights suggest a clear division of labor for researchers employing language models in hypothesis generation. Models are good at identifying local gaps and extending researcher-defined trajectories, producing outputs that reside farther from existing literature than typical published papers, a distance that increases when peripheral papers are used as context. However, they show lower directional novelty than human papers, and standard prompt adjustments fail to overcome this limitation. Our results suggest that researchers should set the direction of inquiry and use models for detailed exploration within established frameworks while rigorously distinguishing between semantic distance and true directional novelty in generated ideas.

Three practical consequences follow. First, evaluation metrics that reward novelty based solely on semantic distance are easily confounded; judges instructed to demand conceptual departures increase distance without improving directional novelty. Systems relying on distance-based metrics may overstate their originality and require directional evaluation standards. Second, directional novelty can be quantified efficiently using embedding-based metrics such as ONS, which requires only hypothesis and literature embeddings to flag outputs remaining within established paradigms for pruning or human review. Low directional scores do not render hypotheses invalid, but they clarify their structural position relative to existing literature. Third, in our experiments neither larger models nor iterative refinement removed the directional deficit. While automated pipelines operate at speeds unmatched by human teams, our results give no reason yet to hand them the choice of research direction.

Our findings also carry notable implications for science policy. Expectations that language models will accelerate discovery through autonomous ideation warrant reevaluation, given that the directional contraction persists across fields, models and most tested configurations. Although extending established directions has intrinsic scientific value, as high-impact papers combine mostly conventional prior work with a small intrusion of atypical combinations~\citep{uzzi2013atypical}, scaling automated generation within established paradigms risks exacerbating the narrowing effect of artificial intelligence on scientific exploration~\citep{hao2026artificial}. To the extent that ONS captures directional novelty, and within the configurations we tested, funding agencies and research institutions should treat language models as amplifiers of human-led research programs rather than autonomous discovery agents, encourage reporting of directional metrics alongside distance measures in AI evaluations, and continue to support exploratory, human-directed scientific inquiry.

Our study has limitations. We assess hypotheses only in terms of novelty, but in practice useful hypotheses must also be falsifiable~\citep{popper2005logic} and experimentally feasible, which we do not evaluate. Distance and direction are quantified within the SPECTER2 embedding space and serve as proxies for novelty. We did not test sensitivity to the encoder, embedding normalization, abstract length, or information density, and ONS has not been validated against historical examples of new research directions. ONS responds to controlled changes in content (Supplementary Information), but by design it does not reward new combinations of established directions. It isolates a narrower property of introducing directions absent from the literature, which is what we mean by a new research direction~\citep{kuhn1970structure}. Validation against expert judgment also remains limited. Pseudo papers are generated from a controlled focal set, but human-authored comparator papers were not written with the same focal set in mind. We also test a subset of models and topics, and deployed systems incorporating tool use, external literature search or iterative experimentation may exhibit distinct behaviors. Our context-selection strategies are not literature-retrieval systems. General-purpose language models constitute the scope of this study, whereas specialized data-driven architectures such as AlphaFold~\citep{jumper2021highly} and Evo 2~\citep{brixi2026genome} fall outside our scope. Nonetheless, across the tested configurations, language models function primarily to extend established scientific trajectories. Advancing directional novelty in artificial intelligence may require systems incorporating external inputs absent from training corpora, including empirical data, experimental results, and human judgment, rather than the enhanced recombination of existing literature. While language models can automate scientific workflows at scale, our results suggest that they cannot yet produce new research directions as reliably as human researchers.

\section*{Methods}
\subsection*{Dataset Construction}
We use the OpenAlex dataset~\citep{priem2022openalex, openalex}, an open scholarly database succeeding the Microsoft Academic Graph (MAG). As of October 2026, OpenAlex contains 517 million research papers, together with associated metadata such as citation records. From this corpus, we select English research papers published in journals and conferences, and exclude those with incomplete titles or abstracts. Following~\cite{hao2026artificial}, we study six natural-science fields: biology, chemistry, geology, materials science, medicine and physics. They map onto OpenAlex's hierarchy of papers as follows.

\begin{itemize}
    \item \textbf{Physics}: \textit{Physical Sciences}  (OpenAlex domain) $\rightarrow$ \textit{Physics and Astronomy}  (OpenAlex field)
    \item \textbf{Biology}: \textit{Life Sciences} $\rightarrow$ \textit{Biochemistry, Genetics and Molecular Biology}
    \item \textbf{Chemistry}: \textit{Physical Sciences} $\rightarrow$ \textit{Chemistry}
    \item \textbf{Materials Science}: \textit{Physical Sciences} $\rightarrow$ \textit{Materials Science}
    \item \textbf{Geology}: \textit{Physical Sciences} $\rightarrow$ \textit{Earth and Planetary Sciences}
    \item \textbf{Medicine}: \textit{Health Sciences} $\rightarrow$ \textit{Medicine}
\end{itemize}




\subsection*{Experimental Design}

Our experiments emulate hypothesis ideation in LLM-based automated research systems. This process assumes that users pick initial papers as context and ask for research hypotheses based on input papers. Here we did not incorporate external LLM-based literature search tools as those can introduce another layer of bias~\citep{chelli2024hallucination}. We further assume that the ideation process consists of two main components: an \textit{LLM Ideator} for hypothesis generation and an \textit{LLM Judge} for hypothesis review. This multi-turn dialogue between ideation and judgment to refine ideas is a common pipeline within current state-of-the-art autoresearch systems that aims to produce high-quality research hypotheses progressively. 

\paragraph{LLM Ideator}
The role of this LLM is to generate scientific ideas. These ideas are in the form of a \textit{title} and an \textit{abstract} of a hypothetical research paper, which we call a \textbf{pseudo paper}. It takes two inputs: selected papers and a prompt.

\paragraph{LLM Judge}
The role of this LLM is to judge the hypotheses generated by the LLM ideator. Its input is a batch of LLM-generated hypotheses. The output is feedback based on the input hypotheses and predefined criteria. 



\paragraph{Models}

Our analysis includes closed-source and open-source models listed below:
\begin{itemize}
    \item Closed-source
    \begin{itemize}
        \item \textbf{Gemini}: Gemini-3.1-Flash-Lite, Gemini-3.8-Flash
        \item \textbf{GPT}: GPT-5.4, GPT-5.4-Nano
    \end{itemize}
    \item Open-source
    \begin{itemize}
        \item \textbf{Qwen}: Qwen3.5-9B, Qwen3.5-35B-A3B
        \item \textbf{DeepSeek}: DeepSeek-V4-Flash, DeepSeek-V4-Pro
    \end{itemize}
\end{itemize}




\paragraph{LLM Personas}
Given existing evidence that role-playing affects LLM outputs~\citep{kong2024better}, and the use of roles in AI for science systems~\citep{lu2026towards}, we experiment with three types of LLM personas when using the Ideator and the Judge: \textit{neutral}, \textit{radical} and \textit{conservative}. We provide prompt templates for the three personas in the Supplementary Information.

\subsection*{Measures of Distance and Direction in Embedding Space}
Following previous work~\citep{hao2026artificial}, papers are embedded into 768-dimensional space using the SPECTER2 model~\citep{singh2023scirepeval}. Within the embedding space, we use three measures of novelty. The first two measures are based on Euclidean distance and quantify the distance between a hypothesis paper and a topic. The third measure is defined for a group of hypothesis papers, and quantifies the alignment between the hypotheses and a set of reference papers in the embedding space. Here, a topic is represented as a finite set of embedding vectors representing a specific scientific category. These measures are always computed within the original 768-dimensional space.

\paragraph{Distance to topic centroid (DTC)}
measures the distance between a hypothesis paper $\mathrm{p}^*$ and the centroid of a given OpenAlex topic $\mathcal{T}$, computed from a reference set $\mathcal{R} = \{ \mathrm{p}[i]: i = 1,\cdots,n\}$ of papers from $\mathcal{T}$ (see \textit{Paper sets}). It is defined as
\begin{equation}
    \text{DTC} = \norm{\mathrm{p}^*-\mathrm{c}_{\mathcal{R}}}_2,
\end{equation}
where the centroid is the sample average over the reference set.
\begin{equation}
\label{eq:reference-centroid}
    \mathrm{c}_{\mathcal{R}} = \frac{1}{n}\sum^{n}_{i=1}\mathrm{p}[i].
\end{equation}
DTC captures how far a hypothesis paper is from an ``average paper'' within the topic $\mathcal{T}$.

\paragraph{Distance to nearest seed paper (DNS)}
measures the distance between a hypothesis paper $\mathrm{p}^*$ and its nearest neighbor within a reference set $\mathcal{R} = \{ \mathrm{p}[i]: i = 1,\cdots,n\}$. For DNS, the reference set is the set of seed papers in the same topic. It is defined by
\begin{equation}
    \text{DNS} = \min_{p[i] \in \mathcal{R}}\norm{\mathrm{p}^*-\mathrm{p}[i]}_2
\end{equation}
Compared to the centroid measure, DNS measures the distance to the nearest paper overall. A high DNS indicates that a hypothesis paper is not near any paper in the topic. A high DNS with an average DTC indicates that the hypothesis paper occupies a cavity or a void in the topic cluster.

\paragraph{Orthogonal novelty score (ONS)}
measures how far a paper $\mathrm{p}^*$ points outside the directions spanned by a set of reference papers $\mathcal{R} = \{ \mathrm{p}[i]: i = 1,\cdots,n\}$, which is typically a subset of a given topic. ONS is calculated in two steps: first, a linear subspace of $\mathbb{R}^{768}$ is constructed using $\mathcal{R}$, and second, ONS measures how much of $\mathrm{p}^*$ is explained by its orthogonal complement to this subspace. 

Formally, we proceed as follows. Given the reference set $\mathcal{R}$, we define the reference centroid $\mathrm{c}_\mathcal{R}$ as in \eqref{eq:reference-centroid}. PCA is applied to the centered reference set, $\{ \mathrm{p}[i] - \mathrm{c}_\mathcal{R}\}$, to obtain an orthonormal basis, which we express as a matrix $\mathrm{V}_\mathcal{R} \in \mathbb{R}^{(n-1) \times 768}$. Because we use centered PCA analysis, the reference matrix $\mathrm{V}_\mathcal{R}$ has rank at most $\min(n-1,768)$. Since we retain all PCA components, and since collinearity in the SPECTER2 embedding space is extremely rare, the maximum rank is usually achieved. The row space of $\mathrm{V}_\mathcal{R}$ defines a linear subspace of $\mathbb{R}^{768}$ representing the semantic directions explored by the reference set.

Next, we consider the centered embedding vector, $\mathrm{p}^*_c = \mathrm{p}^* - \mathrm{c}_\mathcal{R}$. The part of this vector aligned with the reference subspace is given by 
$$
\mathrm{p}^*_{||} = \mathrm{p}^*_c \mathrm{V}_\mathcal{R}^\top  \mathrm{V}_\mathcal{R} 
$$
and the orthogonal part is given by 
$$\mathrm{p}_\perp^* = \mathrm{p}^*_c - \mathrm{p}^*_{||}.$$ 
ONS is defined to be the fraction of the squared magnitude that is orthogonal to the reference subspace, 
\begin{equation}
    \label{eq:ONS}
    \text{ONS}(\mathrm{p}^*,\mathcal{R})
    =
    \frac{\|\mathrm{p}^*_\perp\|^2}{\|\mathrm{p}^*_c\|^2}
    =
    \frac{\| (\mathrm{p}^* - \mathrm{c}_\mathcal{R})( \mathrm{I} - \mathrm{V}^\top_\mathcal{R} \mathrm{V}_\mathcal{R})\|^2}{\| \mathrm{p}^* - \mathrm{c}_\mathcal{R} \|^2},
\end{equation}
where $\mathrm{I}$ is the identity matrix in $\mathbb{R}^{768}$.

By the Pythagorean theorem, ONS takes values between 0 and 1. A paper with an ONS near 1 explores a completely orthogonal semantic direction compared to the reference set. On the other hand, a hypothesis paper with an ONS of 0 is completely explained by the reference set, indicating that the paper's hypothesis is contained within the conceptual directions of the reference set. 

ONS is related to the cosine similarity measures used in other works~\citep{si2025can}. In particular,
\begin{equation}
    \label{eq:cosine}
    \text{ONS}(\mathrm{p}^*,\mathcal{R})
    =
    1 - \frac{\|\mathrm{p}^*_{||}\|^2}{\|\mathrm{p}^*_c\|^2}
    =
    1 - \cos\left(\mathrm{p}^*_c, \mathrm{p}^*_{||}\right)^2,
\end{equation}
where $\cos(\cdot, \cdot)$ is the cosine similarity of two centered vectors. This formulation diverges from existing uses of cosine similarity because $\mathrm{p}^*_c$ is compared with the projection $\mathrm{p}^*_{||}$, as opposed to the reference papers $\mathrm{p}[i]$. This makes ONS more conservative than traditional cosine similarity, because the parallel projection $\mathrm{p}^*_{||}$ is the linear combination of reference papers that is nearest to $\mathrm{p}^*_c$. Hence,
\begin{equation}
    \label{eq:ONS_inequality}
    \cos\left(\mathrm{p}^*_c, \mathrm{p}^*_{||}\right) \geq \max_{\mathrm{p}[i] \in \mathcal{R}} \cos\left(\mathrm{p}^*_c, \mathrm{p}_c[i]\right),
\end{equation}
where $\mathrm{p}_c[i]$ is the $i$-th centered reference paper. Since ONS differs from cosine similarity by a decreasing function $1-x^2$, the inequality \eqref{eq:ONS_inequality} implies that our proposed ONS measure is lower than the corresponding per-reference-paper scores. ONS therefore rates directional novelty lower than its per-reference-paper counterparts. For uncentered vectors in the SPECTER2 embedding space, there is a strong empirical correlation between the maximum cosine similarity and the Euclidean distance to the nearest paper, shown in the Supplementary Information. 

ONS offers a clear advantage over per-reference-paper cosine similarity. Rather than comparing a hypothesis against individual papers, ONS measures the fraction of a hypothesis's conceptual variance that falls outside the entire subspace spanned by the reference literature. This quantity does not depend on which specific reference paper is chosen for comparison, making it a more reliable summary of directional novelty. Please refer to Supplementary Information for technical details and results of ONS validation and simulation studies. 

\subsection*{Paper sets} 
Here we clarify terminology used repeatedly throughout our study:
\begin{itemize}
    \item \textit{Pseudo papers} consist of a title and abstract generated by the LLM ideator. 
    \item \textit{Focal papers} are OpenAlex papers that are provided to the LLM ideator as context. 
    \item \textit{Reference papers} are OpenAlex papers that are used to define the ONS subspace. In all main analyses, the focal set and the reference set are disjoint.
    \item \textit{Comparator papers} are OpenAlex papers that we compare to the pseudo papers of the same topic, using the proposed measures of distance and direction. These papers are also not included in the focal set.
\end{itemize}

\subsection*{Statistical analysis}
All comparisons are made within topics. For each topic and metric we compute the win rate $P(\text{pseudo} > \text{comparator})$ over all pseudo--comparator pairs, and the ratio of the median pseudo paper to the median comparator paper. Topic-level values are summarized by their median within each field, and fields are weighted equally by taking the median of the field medians, so that fields with many topics (for example Medicine, which contributes 42 of the 86 topics in the intervention experiments) do not dominate. Confidence intervals (95\%) and $P$ values come from a bootstrap over topics, so individual hypotheses are not treated as independent units. Comparisons between interventions, and between dialogue turns, are paired within topic. Each pseudo paper is a distinct generated sample, whereas the same set of held-out comparator papers is reused across experimental conditions within a topic.

Each pooled value has 10,000 bootstrap replicates. In each replicate, topics are drawn with replacement within each field (within each subfield for the field-scale analysis, Fig.~\ref{fig:2}). The pooled value is then recomputed as above. Fields and subfields are never resampled, so the intervals describe how precisely the median of each field is known. They do not describe how much the result would vary across other fields. $P$ values are two-sided and come from the replicates: $P = 2 \times min(L + 1, H + 1)/(B + 1)$, capped at 1. Here $B$ is the number of replicates, and $L$ and $H$ are the numbers of replicates at or below and at or above the null value. The null value is 0.5 for a win rate, 1 for a ratio of medians and 0 for a difference. The smallest $P$ this can give is 0.0002, which we report as $P < 0.001$. The 95\% intervals in the figures are the 2.5th and 97.5th percentiles of 2,000 replicates (seed 0), and a filled marker (used in the Supplementary Information) means the interval excludes the null value. Some comparisons contrast two quantities measured on the same topics, such as two models or two metrics. For these, both pooled values are computed from the same resampled topics in each replicate, and the test is on their difference. Other comparisons contrast two groups of hypotheses within a topic, such as a later dialogue turn against the first. For these, we compute the win rate of one group over the other within each topic and pool it as above, with null value 0.5. $P$ values are not adjusted for multiple comparisons. We base conclusions on effects that are consistent across conditions, models and fields rather than on any single test, and we report the number of significant cells alongside the number tested.

\section*{Data availability}
The OpenAlex dataset used in this study is publicly available at~\url{https://openalex.org/}. The generated source data underlying the analyses and figures are available at~\url{https://github.com/CHAI-UK/LLM-Hypothesis-Generation-Study}.

\section*{Code availability}
The code to replicate the results presented in the Article and Supplementary Information is available via GitHub at~\url{https://github.com/CHAI-UK/LLM-Hypothesis-Generation-Study} and is archived on Zenodo at~\url{https://doi.org/10.5281/zenodo.24756238}. The SPECTER2 text embedding model is available at~\url{https://huggingface.co/allenai/specter2}.
We also provide a visual demonstration of the study, available at:~\url{https://chai-uk.github.io/arcadia-project-page/}.


\bibliography{references}
\bibliographystyle{tmlr}


\section*{Acknowledgments}
This work was supported by the Engineering and Physical Sciences Research Council (EPSRC) through the Causality in Healthcare AI Hub (CHAI), grant number EP/Y028856/1.

\section*{Author contributions}
Z.L., K.B., D.M., and S.A.T. conceptualized and designed the study. Z.L. and D.B. preprocessed the data and wrote the codebase for the experiments. Z.L., K.B., and D.M. created and analyzed figures, and drafted the first version of the manuscript. S.A.T. oversaw the project. All authors reviewed and revised the final manuscript.

\section*{Competing interests}
The authors declare no competing interests.


\ifdefined\CombinedVersion
    \clearpage

    \newcommand{\SupplementaryEmbedded}{}
\ifdefined\SupplementaryEmbedded
    \let\supplementaryend\relax
\else
    \documentclass[10pt]{article}
    \let\SupplementaryAddContentsLine\addcontentsline
    \usepackage[preprint]{tmlr}
    \let\addcontentsline\SupplementaryAddContentsLine
    \usepackage{url}
    \usepackage{graphicx}
    \usepackage{amsmath}
    \usepackage{amssymb}
    \usepackage{amsthm}
    \usepackage{mathtools}
    \usepackage{booktabs}
    \usepackage[table]{xcolor}
    \usepackage{longtable,array}
    \usepackage{fvextra}
    \usepackage{needspace}
    \usepackage[most]{tcolorbox}
    \usepackage[hidelinks]{hyperref}
    \hypersetup{pdftitle={Supplementary Information}}
    \DeclarePairedDelimiter{\norm}{\lVert}{\rVert}
    \newcommand{\supplementaryend}{%
        \clearpage
        \phantomsection
        \addcontentsline{toc}{section}{References}
        \bibliographystyle{tmlr}
        \bibliography{references}
        \end{document}%
    }
    \begin{document}
    
\fi

\renewcommand{\contentsname}{Table of Contents}
\begin{center}
        {\LARGE\bfseries\sffamily Supplementary Information\par}
    \end{center}
    \vspace{1em}
    \setcounter{tocdepth}{1}
    \tableofcontents
    \clearpage

\definecolor{prompttitlebg}{HTML}{E9EFF4}

\newcommand{\supplementarysection}[1]{%
    \clearpage
    \subsection*{#1}%
        \addcontentsline{toc}{section}{#1}%
}
\newcommand{\maintextfigure}[1]{%
    \ifdefined\SupplementaryEmbedded
        Figure~\ref{fig:#1}%
    \else
        Figure~#1 in the main text%
    \fi
}

\supplementarysection{OpenAlex categorical hierarchy}

OpenAlex~\citep{priem2022openalex,openalex} organizes scientific works using a hierarchical classification of domains, fields, subfields, and topics: 
\begin{equation}
    \label{eq:openalex-hierarchy}
\text{domain} \supset \text{field} \supset \text{sub-field} \supset \text{topic} \ni \text{paper}.
\end{equation}
Our analyses use six scientific fields, which we refer to using simplified study labels following \citet{hao2026artificial}. Table~\ref{tab:openalex_hierarchy} shows the mapping between these labels and the corresponding OpenAlex hierarchy. Individual research topics used in our experiments are nested within these fields through the OpenAlex classification.

\begin{table}[htbp]
\centering
\caption{Mapping between study field labels and the corresponding OpenAlex hierarchy.}
\label{tab:openalex_hierarchy}
\small
\renewcommand{\arraystretch}{1.25}
\vspace{6pt}
\begin{tabular}{@{}>{\raggedright\arraybackslash}p{0.20\linewidth}>{\raggedright\arraybackslash}p{0.23\linewidth}>{\raggedright\arraybackslash}p{\dimexpr0.57\linewidth-4\tabcolsep\relax}@{}}
\toprule
\rowcolor{prompttitlebg}
{\sffamily\bfseries Study label} & {\sffamily\bfseries OpenAlex domain} & {\sffamily\bfseries OpenAlex field} \\
\midrule
Biology & Life Sciences & Biochemistry, Genetics and Molecular Biology \\
\addlinespace[2pt]
Physics & Physical Sciences & Physics and Astronomy \\
\addlinespace[2pt]
Chemistry & Physical Sciences & Chemistry \\
\addlinespace[2pt]
Materials Science & Physical Sciences & Materials Science \\
\addlinespace[2pt]
Geology & Physical Sciences & Earth and Planetary Sciences \\
\addlinespace[2pt]
Medicine & Health Sciences & Medicine \\
\addlinespace[2pt]
\bottomrule
\end{tabular}
\end{table}

The experiments discussed in the main text use the full corpus for analysis. For some of the additional experiments discussed in the supplemental material, we also selected topics from each study field to focus on. For reference, these topics are listed in Table~\ref{tab:largest_topics}.

\begin{table}[htbp]
\centering
\caption{Selected topics from each field that contain the largest number of papers among all topics in the same field. The original OpenAlex topic IDs and topic names are also provided.}
\label{tab:largest_topics}
\small
\renewcommand{\arraystretch}{1.25}
\vspace{6pt}
\begin{tabular}{@{}>{\raggedright\arraybackslash}p{0.19\linewidth}>{\raggedright\arraybackslash}p{0.21\linewidth}>{\raggedright\arraybackslash}p{0.12\linewidth}>{\raggedright\arraybackslash}p{\dimexpr0.48\linewidth-6\tabcolsep\relax}@{}}
\toprule
\rowcolor{prompttitlebg}
{\sffamily\bfseries Field study label} & {\sffamily\bfseries Topic study label} & {\sffamily\bfseries Topic ID} & {\sffamily\bfseries OpenAlex topic name} \\
\midrule
Biology & Biosensing & T10207 & Advanced biosensing and bioanalysis techniques \\
\addlinespace[2pt]
Physics & Particle Physics & T10048 & Particle physics theoretical and experimental studies \\
\addlinespace[2pt]
Chemistry & MOF & T10096 & Metal-Organic Frameworks: Synthesis and Applications \\
\addlinespace[2pt]
Materials Science & Catalytic Processes & T10192 & Catalytic Processes in Materials Science \\
\addlinespace[2pt]
Geology & Remote Sensing & T13890 & Remote Sensing and Land Use \\
\addlinespace[2pt]
Medicine & Liver Disease & T10351 & Liver Disease Diagnosis and Treatment \\
\addlinespace[2pt]
\bottomrule
\end{tabular}
\end{table}

\begin{figure}[hp]
    \centering
    \includegraphics[width=0.98\linewidth]{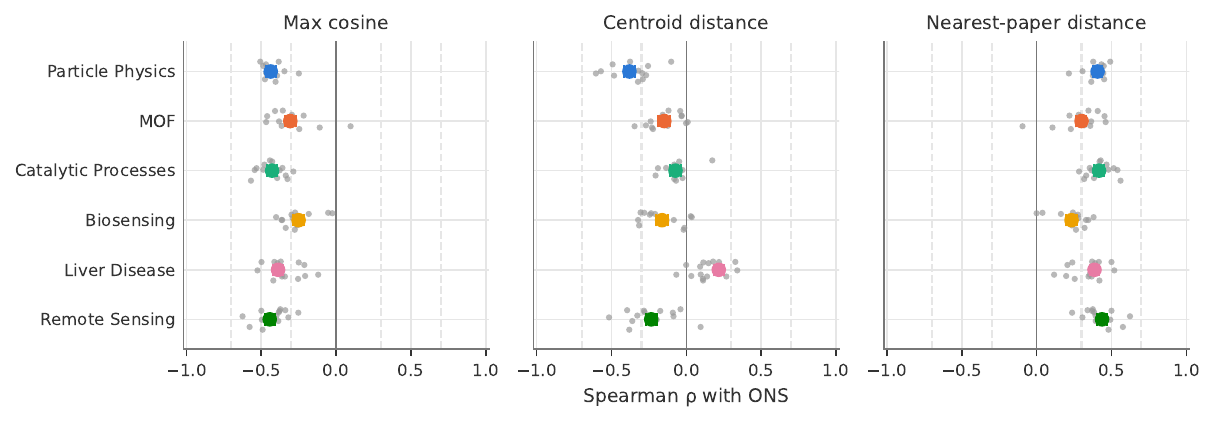}
    \caption{\textbf{ONS is distinct from conventional embedding-space proximity measures.} Spearman correlations between ONS and maximum cosine similarity, distance to the reference centroid, and distance to the nearest reference paper are computed within topic-model runs. We compare metrics across the six topics in Table~\ref{tab:largest_topics}. ONS shows weak-to-moderate associations with maximum cosine similarity and nearest-paper distance, and a weaker association with centroid distance. Points show individual topic-model runs; summary markers show the median correlation across runs.}
    \label{fig:ons-vs-cosine}
\end{figure}

\supplementarysection{Metric redundancy analysis}
To assess whether ONS is redundant with conventional embedding-space proximity measures, we computed Spearman correlations between ONS and maximum cosine similarity, distance to the reference centroid, and distance to the nearest reference paper. Because metric distributions differ across topics, correlations were calculated for each topic rather than only after pooling all hypotheses. ONS showed weak-to-moderate associations with maximum cosine similarity and nearest-paper distance, with median correlations of \(\rho=-0.38\) and \(\rho=0.37\), respectively. Its association with centroid distance was weaker and varied across topics (median \(\rho=-0.11\)). These results indicate that ONS is related to, but not reducible to, conventional measures of semantic proximity such as maximum cosine similarity. Per-topic correlations are shown in Figure~\ref{fig:ons-vs-cosine}.

\begin{figure}[hp]
    \centering
    \includegraphics[width=0.8\linewidth]{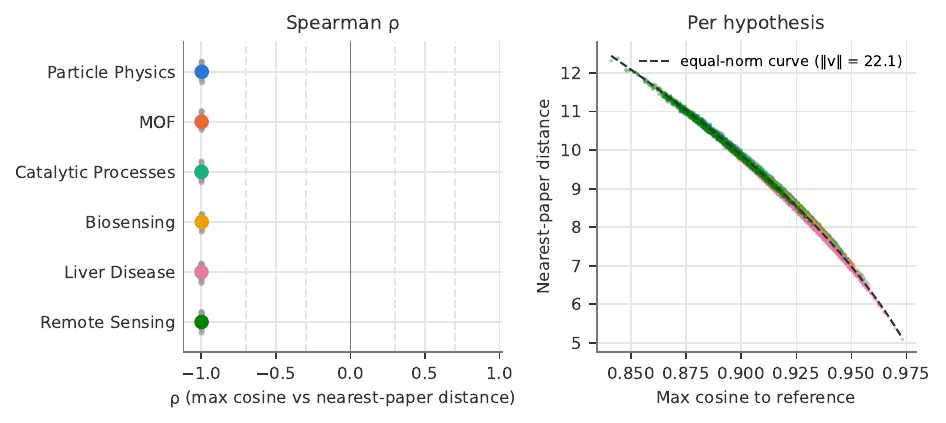}
    \caption{\textbf{Maximum cosine similarity is nearly rank-equivalent to nearest-paper distance metrics.} We plot the maximum cosine similarity versus Euclidean nearest-paper distance to the same 120-paper reference set. The analysis is repeated for each of the six topics in Table~\ref{tab:largest_topics}. Left: Spearman $\rho$ per topic (colored, 95-percent bootstrap CI), and per model run (gray). Right: one point per paper; the dashed curve is $y = 22.1 \sqrt{2-2x}$, the exact relation under the hypothesis that all embedding vectors $\mathrm{v}$ have equal norm, $\|\mathrm{v}\| = 22.1$.}
    \label{fig:cosine-vs-distance}
\end{figure}

\newpage
As shown in Figure ~\ref{fig:cosine-vs-distance}, we also found that maximum cosine similarity and nearest-paper distance were themselves almost perfectly inversely correlated within topics (\(\rho=-0.998\)). This is partially expected because SPECTER2 embedding norms vary little across papers, making Euclidean nearest-neighbor distance and maximum cosine similarity nearly equivalent rankings in this dataset.  We therefore treat these measures as closely related comparators rather than independent tests.

\supplementarysection{Isotropic null hypothesis for ONS}
We characterize ONS under the null hypothesis that embedding vectors have no preferred spatial structure. We present the mathematical derivation of the expected ONS under the null hypothesis in this section for comparison with our results in the main text. 

We first assume a fixed reference set $\mathcal{R}$ with $n$ reference papers. Without a loss of generality, we may orient our coordinate system such that the centroid $\mathrm{c}_\mathcal{R}=0$ and  $\mathrm{V}_\mathcal{R}=[\mathrm{e}_1^\top \cdots \mathrm{e}_{n-1}^\top]^\top$, where $\mathrm{e}_i$ is the $i$-th standard basis vector in $\mathbb{R}^{768}$. We then consider a random vector $\mathrm{u} \in \mathbb{R}^{768}$ with an isotropic (rotation-invariant) probability distribution. Since ONS is invariant to changes in distance to the reference centroid, we may assume $\mathrm{u} \sim \mathcal{N}(0,\mathrm{I})$. The ONS of $\mathrm{u}$ may then be computed
$$
\text{ONS}(\mathrm{u},\mathcal{R})
=
\frac{\sum_{i=n}^{768} u_i^2}{\sum_{i=1}^{768} u_i^2}.
$$
Since each $u_i \sim\mathcal{N}(0,1)$ in this coordinate system, $\sum_{i=s}^r u_i^2 \sim \chi^2(r-s+1)$, and the ratio above is Beta distributed.
This analysis yields the expected value
\begin{equation}
\label{eq:null-hypothesis}
U_n = E[\text{ONS}] =1 - \frac{\min(n-1,768)}{768},
\end{equation}
where the minimum arises from the case that $n$ exceeds the dimension of the embedding space. 
ONS values substantially below $U_n$ indicate that papers are more aligned with the reference set than would be expected from an isotropic embedding. 

For a reference set of $n=20$ papers, this value is $U_{20}=0.975$. For the $n=200$ reference sets used in the main analysis, $U_{200}=1-199/768\approx0.741$.
Across most experiments, we observe ONS values well below this threshold, indicating that embeddings of pseudo papers exhibit spatial structure beyond that expected under the isotropic null model. We show this explicitly in Fig.~\ref{fig:ons-sweep-over-n}, as part of another experiment in which we observe the sensitivity of ONS to $n$.

 \supplementarysection{Validation of ONS against lexical similarity}
ONS measures the directional alignment between a paper and a reference set of papers in a embedding space. If ONS tracks topic direction, it should at least respond to controlled changes in a document's similarity to the reference set. Here we compare ONS to a synthetic benchmark metric which we define as the sentence-mix fraction (SMF) which mixes off-topic papers with papers from the reference set to produce titles and abstracts which has intermediate degrees of relevance to the reference topic. Notably, this experiment operates in text space: mixed abstracts are encoded independently with SPECTER2 before ONS is calculated. We do not assume that sentence mixing produces a corresponding linear combination of paper embeddings. The experiment therefore evaluates whether ONS responds systematically to controlled changes in document content.

Pseudo papers are generated by a sentence-mix algorithm that combines sentences from a source paper $a$ and a target paper $b$ to produce a pseudo paper $u$. We provide a technical description here. The abstracts of both papers are divided into $r$ units (sentences or joined adjacent sentences), where $r$ is the smaller of the two sentence counts. A binary mask $m \in \{0,1\}^r$ determines whether each slot uses a unit from $a$ or $b$. The title is drawn from either $a$ or $b$. The SMF is defined as the word-weighted fraction of text drawn from the target paper:
\begin{equation}
    \text{SMF} = \frac{\sum_{i:m_i=1} w(b_i) + 1_{[t=b]}w(\text{title}_b)}{\sum_{i=1}^n w(u_i) + w(\text{title}_u)},
\end{equation}
where $w(\cdot)$ denotes word count, $u_i$ is the unit in slot $i$ of $u$, and $1_{[t=b]}$ is the indicator function for when the title is drawn from paper $b$. SMF is bounded between 0 and 1, and the extremes 0 and 1 correspond to cases when $u$ is a copy of the source or target paper respectively. 

\begin{figure}[ht]
    \centering
    \includegraphics[width=0.7\linewidth]{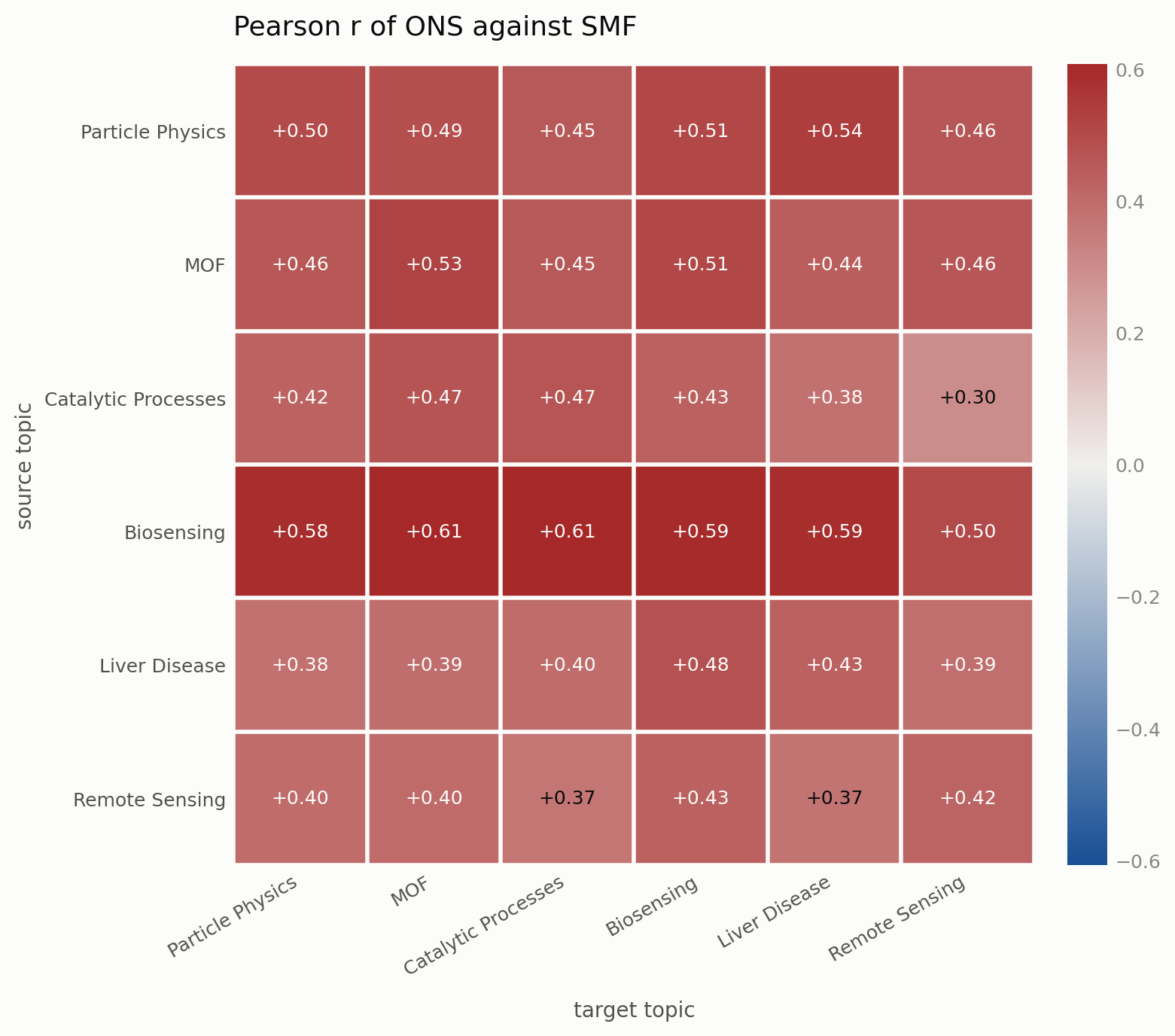}
    \caption{\textbf{ONS correlates with sentence-mix fraction (SMF) across topics.} We mechanistically generate pseudo papers by mixing sentences from a source topic and a target topic, using the six topics in Table~\ref{tab:largest_topics}. The fraction of words in the resulting abstract that come from the target topic is measured as the sentence-mix fraction (SMF). When the reference set for ONS is sampled from the source topic, SMF is a noisy measure of a pseudo paper's relevance to the source topic. The source paper used for sentence mixing is always included in the reference set. The Pearson correlation coefficient for each source target topic pair is extracted and visualized as a heatmap. 
    Note that the Pearson correlation coefficient is invariant to the average value of a distribution, so this plot does not reflect the fact that same-topic source-target pairs show lower ONS overall. 
    }
    \label{fig:ons_heatmap}
\end{figure}

\begin{figure}[ht]
    \centering
    \includegraphics[width=0.9\linewidth]{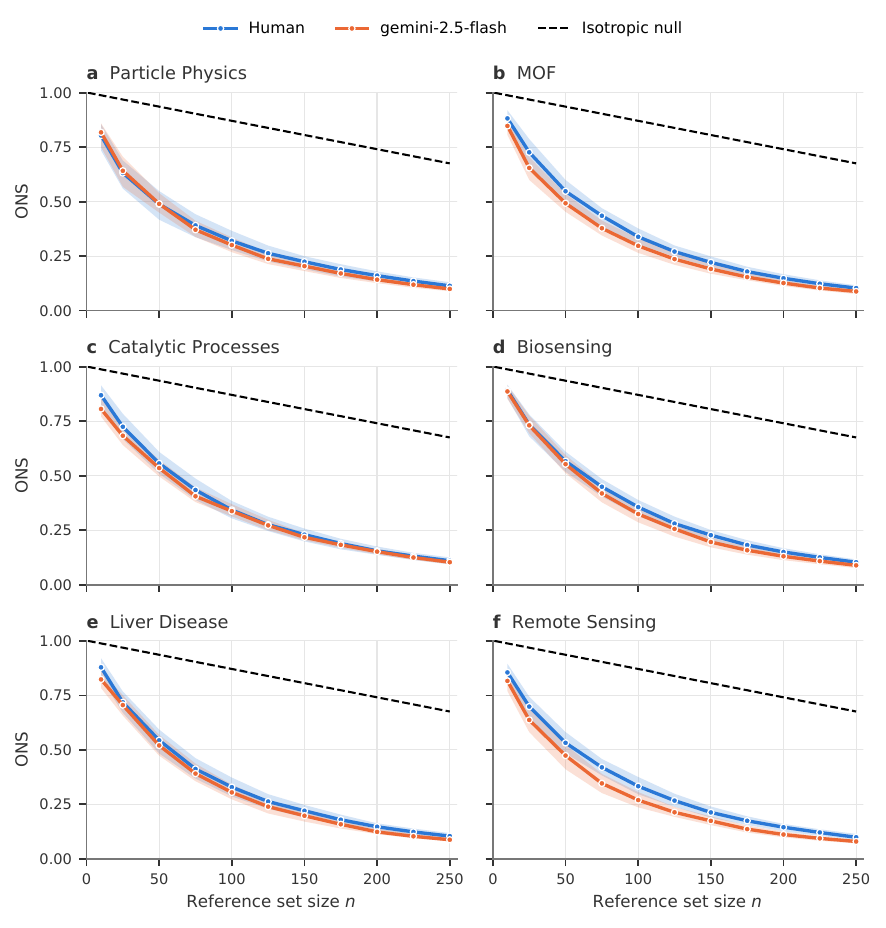}
    \caption{
    \textbf{ONS ranking is consistent across varying reference set sizes.}
    For each of the six topics in Table~\ref{tab:largest_topics}, we compare within-topic ONS of human-authored comparator papers with their LLM-generated counterparts (in this case, the gemini-2.5-flash model). The ONS reference set size is varied by fixing an initial set of 250 reference papers, and keeping only the first $n$ papers in the set. The solid line represents the median ONS and the shaded region represents the interquartile range (25th to 75th percentile). We also show the null hypothesis expected value, $U_n$, as shown in \eqref{eq:null-hypothesis}. The magnitude of ONS is dependent on $n$, but the relative ordering of the two distributions does not depend on $n$. 
    }
    \label{fig:ons-sweep-over-n}
\end{figure}

We performed a small experiment to compare ONS and SMF across the six topics in Table~\ref{tab:largest_topics}. Since some sentences might be domain neutral, we expect SMF to correlate to a moderate degree with ONS, but the relationship cannot be one-one. For each pair of source and target topic, we select 100 pairs of papers for mixing, and produce 10 pseudo papers for each paper pair. To compute ONS, we sample an independent 50-paper reference set for each source topic.
Pearson correlation coefficients for all fields are shown in  Fig. \ref{fig:ons_heatmap} as a heatmap. We find that ONS correlates positively with SMF across all 36 cases (ranging from +0.30 to +0.61). This shows that ONS responds systematically to controlled changes in document content and topic; on its own it does not establish that ONS measures conceptual innovation.


\supplementarysection{Sensitivity of ONS to reference set size}
 The magnitude of ONS naturally decreases as the number of reference papers used to construct the reference subspace increases. To test whether this phenomenon will affect our conclusions in the main text, we consider a small experiment using the six topics in Table~\ref{tab:largest_topics}. 
 
 In Fig. \ref{fig:ons-sweep-over-n}, we visualize how ONS varies with $n$, the size of the reference set. Across all six topics, ONS of both human-authored and LLM-generated papers decrease uniformly as papers are added to the reference set. Despite this effect, human-authored papers consistently score higher or equal to LLM-generated papers.

\supplementarysection{ONS does not change with field size}

\begin{figure}[h]
    \centering
    \includegraphics[width=0.95\linewidth]{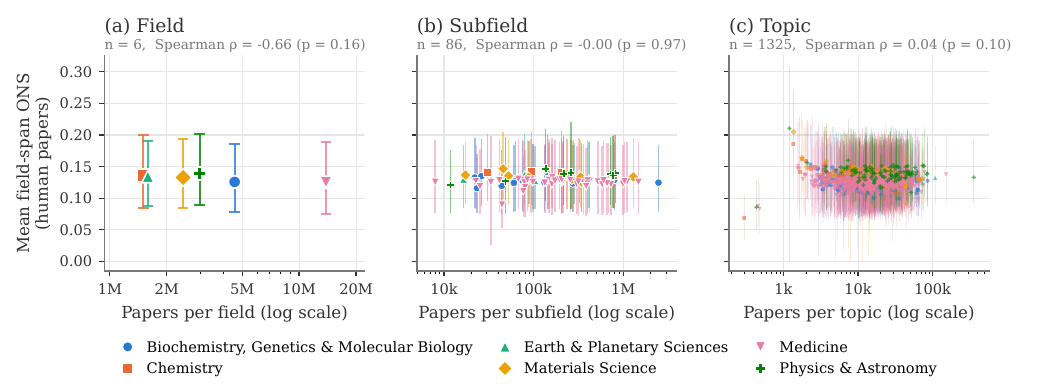}
    \caption{\textbf{ONS remains consistent across OpenAlex categories of different sizes.} Papers in the human-authored corpus are grouped according to their OpenAlex Field, Subfield or Topic. For each paper, ONS is calculated using the field-subspace ONS corresponding to its Field. Points show the mean ONS within each category, and bars show the 5th--95th percentile range. Category size denotes the number of papers assigned to each category. }
    \label{fig:ons-vs-field-size}
\end{figure}

Here we test whether the magnitude of ONS varies significantly with the size of a topic, subfield or field, measured in the number of papers.
In Fig.~\ref{fig:ons-vs-field-size}, we plot the ONS for the corpus of seed papers, dividing the corpus into its constituent fields, subfields and topics according to the hierarchy \eqref{eq:openalex-hierarchy}. We observe no systematic relationship between category size and ONS, with the distribution of category-level estimates remaining broadly consistent across the range of category sizes. This indicates that variation in the number of papers assigned to a field does not substantially affect the measured ONS.

\supplementarysection{Truncated PCA does not change ONS conclusions}
ONS projects each paper onto a PCA subspace determined by a set of reference papers. The main content of the manuscript uses $n=200$ reference papers, with the PCA subspace set to 199 dimensions. We refer to this formulation as \textit{untruncated} ONS.

An alternative approach to defining ONS is to sample $n$ reference papers, then to select some $k\leq\mathrm{rank}(\mathrm{V}_\mathcal{R})$ and compute a $k$-dimensional subspace by retaining only the top $k$ principal components of the reference set. We refer to this variant as \textit{truncated} (and centered) ONS. We show that the ONS deficit between human-authored and LLM-generated papers persists under truncated ONS. Generated hypotheses and the 1000 comparator papers are always projected onto the same reference subspace. For a fixed reference set $\mathcal{R}$, ONS also decreases monotonically as $k$ increases. The inequality between ONS and per-paper cosine similarity derived in Methods holds only for the untruncated subspace; we do not rely on it for truncated ONS.

\begin{figure}[ht]
    \centering
    \includegraphics[width=0.9\linewidth]{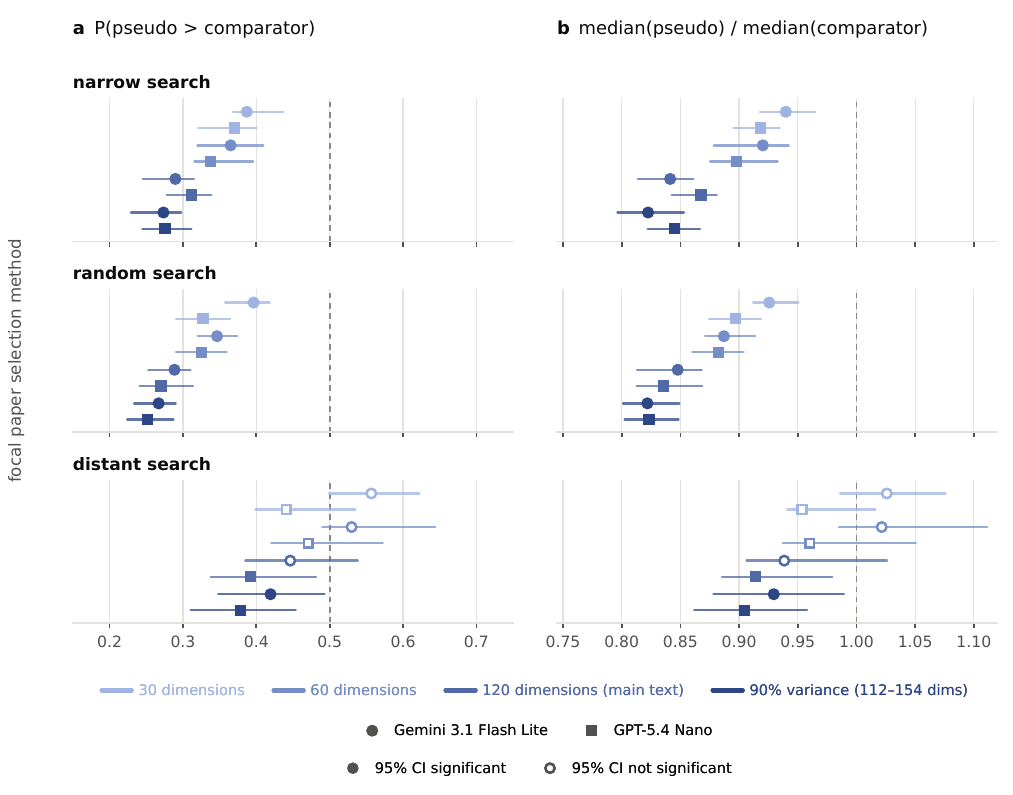}
    \caption{\textbf{ONS deficit is mostly robust to context selection methods.} The main untruncated ONS results (200 papers, 199 dimensions) are recalculated using truncated PCA variants to assess the robustness of ONS and our findings. Here we repeat the analysis of \maintextfigure{4} and vary the number of dimensions $k$. The qualitative pattern remains similar across the tested truncation levels. The variants with the least truncation (112-154 dimensions preserved) correspond to selecting truncation values that preserve the dimensions that holds 90\% of the variance within a topic.}
    \label{fig:supp-fig4-rank}
\end{figure}

\begin{figure}[ht]
    \centering
    \includegraphics[width=0.9\linewidth]{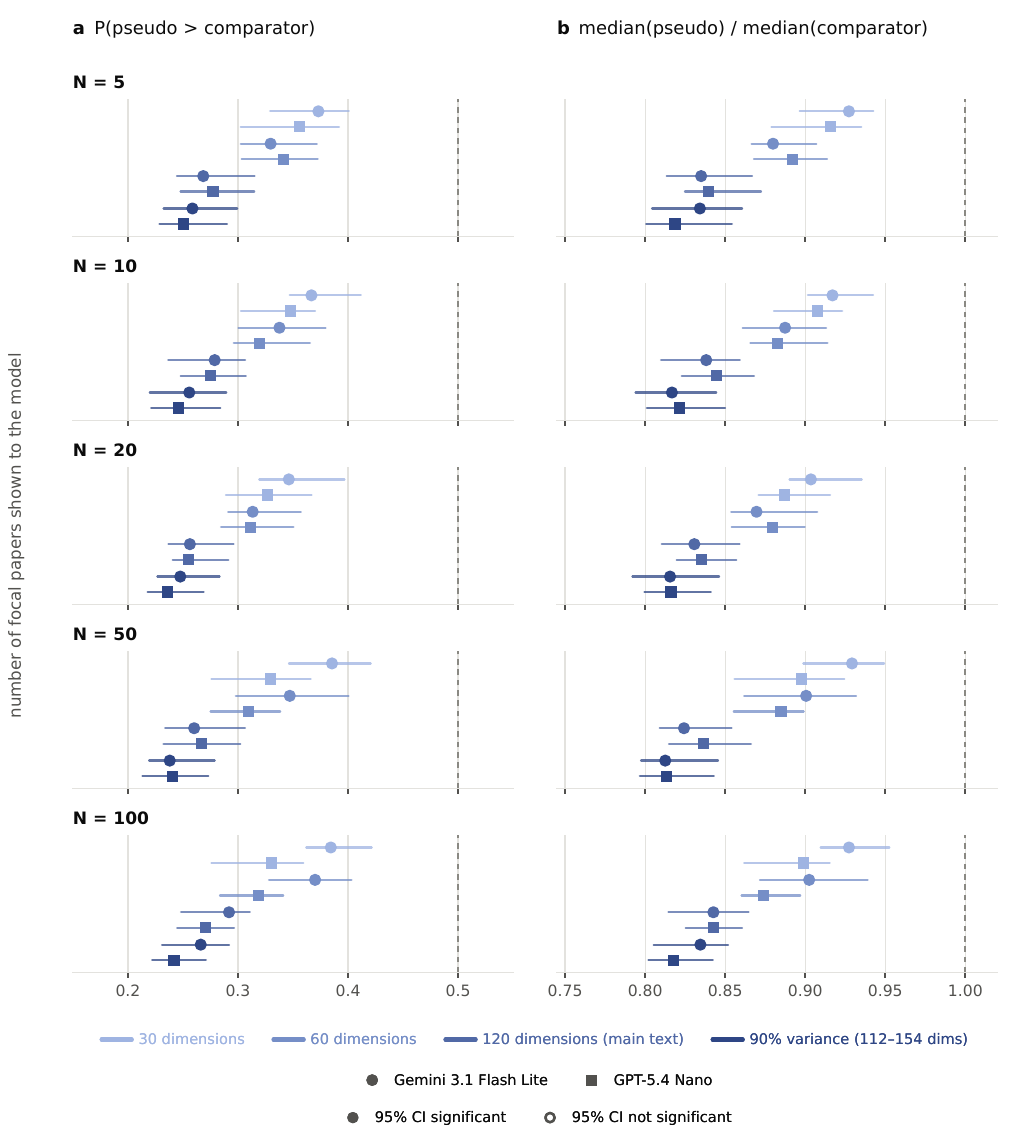}
    \caption{\textbf{Truncated ONS deficit is robust to context size.} Similar to Figure \ref{fig:supp-fig4-rank}, the main untruncated ONS results (200 papers, 199 dimensions) are recalculated using truncated PCA variants to assess the robustness of ONS and our findings. Here, we repeat the analysis of \maintextfigure{3} with truncated variants. The number of chosen dimensions $k$ used to construct the linear subspace are varied along with the number of reference papers $n$.}
    \label{fig:supp-fig3-rank}
\end{figure}

\begin{figure}[ht]
    \centering
    \includegraphics[width=0.9\linewidth]{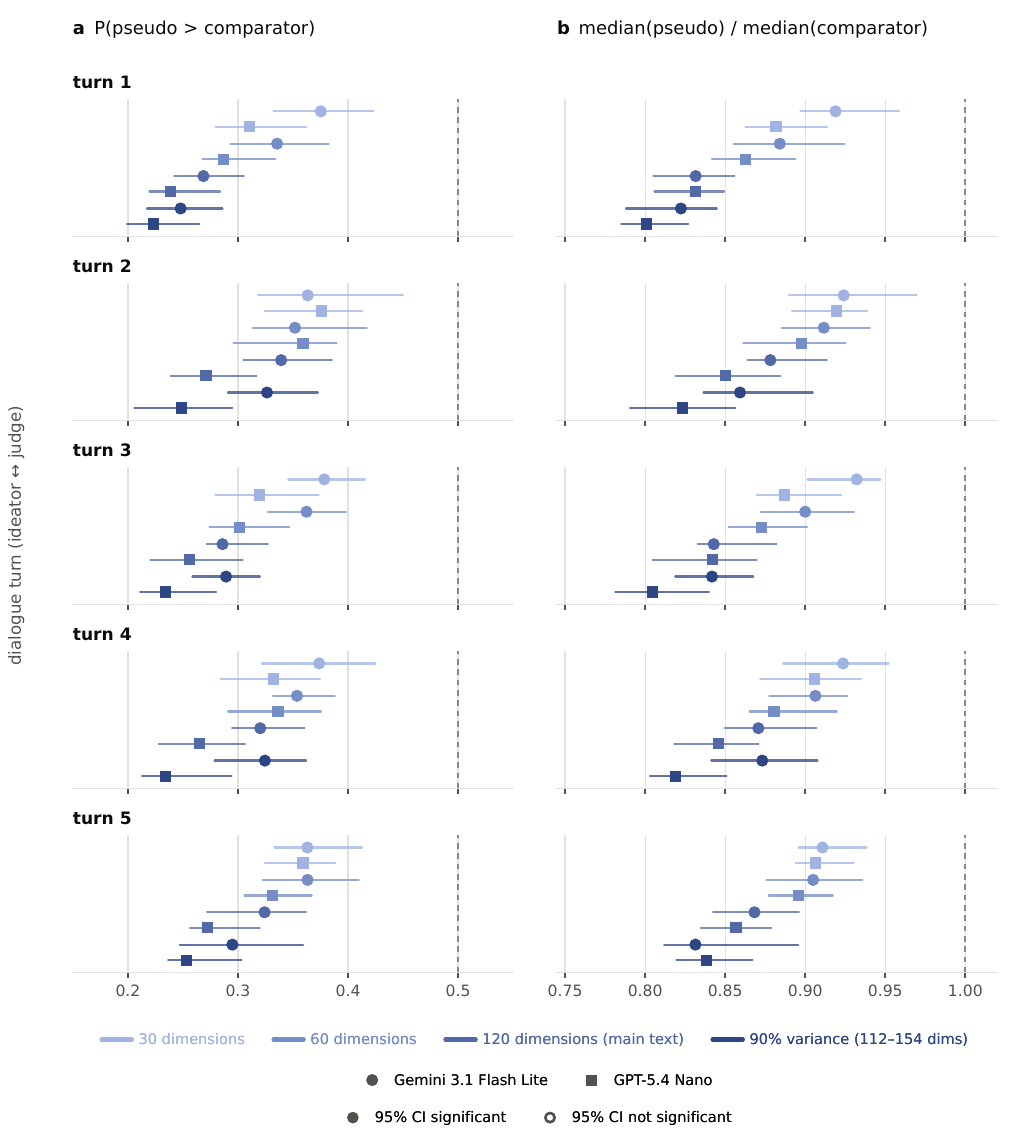}
    \caption{\textbf{Truncated ONS deficit is robust across the LLM feedback.} Similar to Figure \ref{fig:supp-fig4-rank}, the main untruncated ONS results (200 papers, 199 dimensions) are recalculated using truncated PCA variants to assess the robustness of ONS and our findings. Here, we repeat the multiple-round experiment of \maintextfigure{5} using truncated ONS variants. Each subplot shows a fixed round, with the number of dimensions $k$ varied.}
    \label{fig:supp-fig5-rank}
\end{figure}

As presented in Figs~\ref{fig:supp-fig4-rank}-\ref{fig:supp-fig5-rank}, the deficit persists across repeated experiments using truncated ONS.

\newpage

\newpage

\supplementarysection{Prompt Templates}
\begingroup
\setlength{\parindent}{0pt}
\setlength{\parskip}{0.45em}
\setlength{\emergencystretch}{2em}
\definecolor{promptink}{HTML}{263746}
\definecolor{promptrule}{HTML}{CDD7DF}
\definecolor{promptaccent}{HTML}{405F75}
\definecolor{promptbg}{HTML}{F7F9FB}
\definecolor{formatbg}{HTML}{FAFAFA}
\newcounter{promptblock}
\tcbset{promptpanel/.style={
  enhanced jigsaw,breakable,sharp corners,
  colback=promptbg,colframe=promptrule,coltext=promptink,
  colbacktitle=prompttitlebg,coltitle=promptink,
  boxrule=0.4pt,titlerule=0pt,
  borderline west={1.6pt}{0pt}{promptaccent},
  left=4mm,right=4mm,top=3mm,bottom=3mm,
  toptitle=2.5mm,bottomtitle=2.5mm,
  fonttitle=\sffamily\bfseries\small,
  before skip=12pt,after skip=12pt,
  pad at break*=2.5mm}}
\newenvironment{PromptBox}[3]{%
  \Needspace{7\baselineskip}%
  \refstepcounter{promptblock}%
  \begin{tcolorbox}[promptpanel,
    title={Prompt \thepromptblock\quad #1\par\smallskip
      {\normalfont\sffamily\footnotesize #3}},
    title after break={Prompt \thepromptblock\quad #1\enspace\textnormal{(continued)}}]%
  \label{#2}%
}{\end{tcolorbox}}
\newenvironment{FormatBox}[1]{%
  \Needspace{5\baselineskip}%
  \begin{tcolorbox}[promptpanel,colback=formatbg,
    colbacktitle=formatbg,borderline west={1pt}{0pt}{promptrule},
    title={#1},title after break={#1\enspace\textnormal{(continued)}}]%
}{\end{tcolorbox}}
\DefineVerbatimEnvironment{PromptText}{Verbatim}{%
  fontsize=\small,baselinestretch=1.08,
  breaklines=true,breakanywhere=false,
  breaksymbolleft={},breaksymbolright={},
  frame=none,xleftmargin=0pt,xrightmargin=0pt,
  formatcom=\color{promptink},samepage=false}
\newcounter{supppromptsection}
\newcounter{supppromptsubsection}[supppromptsection]
\renewcommand{\thesupppromptsection}{P\arabic{supppromptsection}}
\renewcommand{\thesupppromptsubsection}{\thesupppromptsection.\arabic{supppromptsubsection}}
\newcommand{\supppromptsection}[1]{%
    \refstepcounter{supppromptsection}%
    \subsubsection*{\thesupppromptsection\quad #1}%
    \ifdefined\SupplementaryEmbedded\else
        \addcontentsline{toc}{subsection}{\thesupppromptsection\quad #1}%
    \fi
}
\newcommand{\supppromptsubsection}[1]{%
    \refstepcounter{supppromptsubsection}%
    \paragraph*{\thesupppromptsubsection\quad #1}%
}

\supppromptsection{Prompt for ``LLM Generated Hypotheses Open Fewer New Research Directions''}
\label{supp-prompt:sec:generation}
Each hypothesis is requested as a paper title and abstract. The system message
in Prompt~\ref{supp-prompt:p:generation-system} defines the generation task, and the user
message in Prompt~\ref{supp-prompt:p:initial-user} supplies the focal papers. For successive
batches, Prompt~\ref{supp-prompt:p:history} adds previously generated titles to the user message.

\paragraph{Runtime variables.}
Angle-bracket placeholders are replaced with values for the current topic or model call.

\begingroup
\small
\renewcommand{\arraystretch}{1.25}
\setlength{\LTleft}{0pt}
\setlength{\LTright}{0pt}
\begin{longtable}{@{}>{\raggedright\arraybackslash}p{0.44\linewidth}>{\raggedright\arraybackslash}p{\dimexpr0.56\linewidth-2\tabcolsep\relax}@{}}
\toprule
\rowcolor{prompttitlebg}
\sffamily\bfseries Variable & \sffamily\bfseries Definition \\
\midrule
\endfirsthead
\toprule
\rowcolor{prompttitlebg}
\sffamily\bfseries Variable & \sffamily\bfseries Definition (continued) \\
\midrule
\endhead
\midrule
\multicolumn{2}{r}{\footnotesize\itshape Continued on the next page} \\
\endfoot
\bottomrule
\endlastfoot
\texttt{<TOPIC>} & Research topic for the current generation task. \\
\addlinespace[2pt]
\texttt{<N\_FOCAL>} & Number of focal papers supplied to the model. \\
\addlinespace[2pt]
\texttt{<BATCH\_SIZE>} & Number of hypotheses requested in one call. Several calls can contribute to the total for a topic. \\
\addlinespace[2pt]
\texttt{<FOCAL\_PAPERS>} & Numbered focal-paper titles and abstracts, using the format below. A blank line separates records. \\
\addlinespace[2pt]
\texttt{<INDEX>} & Consecutive paper number in the focal-paper list. \\
\addlinespace[2pt]
\texttt{<TITLE>}\newline \texttt{<ABSTRACT>} & The paper's title and abstract, respectively. \\
\addlinespace[2pt]
\texttt{<OPTIONAL\_TITLE\_HISTORY>} & Empty before any titles have been generated. Otherwise, contains the block in Prompt~\ref{supp-prompt:p:history}. \\
\addlinespace[2pt]
\texttt{<GENERATED\_TITLE\_LIST>} & Previously generated titles, one per line, each prefixed by a hyphen. \\
\addlinespace[2pt]
\end{longtable}
\endgroup

\begin{FormatBox}{Focal-paper record format}
\begin{PromptText}
[<INDEX>] Title: <TITLE>
Abstract: <ABSTRACT>
\end{PromptText}
\end{FormatBox}

\begin{PromptBox}{Cross-model generation}{supp-prompt:p:generation-system}{System message}
\begin{PromptText}
You are an ambitious research scientist looking to publish work that will contribute significantly to the research topic described below. You have broad knowledge of the topic — its methods, its recent findings, its controversies, and the open questions that define ongoing work. You are creative and rigorous, and you propose ideas that are novel, impactful, and feasible given the state of the field.
The research topic for this session is: <TOPIC>.
Your task is to read a set of seed research papers (titles and abstracts) and generate novel research ideas inspired by them.
Each idea must be expressed as a realistic research paper with a title and an abstract.

Guidelines:
- The generated papers must be NOVEL research ideas, not summaries or paraphrases of the seed papers.
- Draw inspiration from themes, methods, or gaps you observe across the seed set.
- Come up with impactful and creative research ideas. Do not overfit to the specific systems, samples, materials, or interventions in the seed papers — favour ideas with wider significance that could generalise across sub-areas of the topic or bridge it with adjacent disciplines. Be realistic: an idea should be feasible with methods and resources reasonably available to researchers in this area.
- Each abstract should be 100-200 words, written in a formal academic style.
- Return ONLY a valid JSON array. Do not include any explanation or markdown outside the JSON.

Output format (strict JSON array):
[
  {
    "title": "...",
    "abstract": "..."
  },
  ...
]
\end{PromptText}
\end{PromptBox}

\begin{PromptBox}{Generation from focal papers}{supp-prompt:p:initial-user}{User message}
\begin{PromptText}
Below are <N_FOCAL> seed research papers:

<FOCAL_PAPERS>
<OPTIONAL_TITLE_HISTORY>
Based on the seed papers above, generate <BATCH_SIZE> additional novel pseudo-research-paper ideas that are distinct from all titles listed above and have wider significance beyond the specific methods or findings in the seed papers. Return exactly <BATCH_SIZE> items as a JSON array.
\end{PromptText}
\end{PromptBox}

\begin{PromptBox}{Optional title-history block}{supp-prompt:p:history}{Inserted at <OPTIONAL\_TITLE\_HISTORY>; omitted when no titles exist}
\begin{PromptText}

The following titles have already been generated. Do NOT repeat, paraphrase, or produce ideas that overlap with any of these:
<GENERATED_TITLE_LIST>
\end{PromptText}
\end{PromptBox}

\clearpage
\supppromptsection{Prompt for ``Context Moves Hypotheses Outward More Than Into New Directions''}
\label{supp-prompt:sec:context}
The context-size and context-selection experiments use the system message below
with the user message in Prompt~\ref{supp-prompt:p:initial-user} and the optional title history
in Prompt~\ref{supp-prompt:p:history}. The role description and significance instruction
differ from those in Prompt~\ref{supp-prompt:p:generation-system}.

\paragraph{Runtime variables.}
\begingroup
\small
\renewcommand{\arraystretch}{1.25}
\setlength{\LTleft}{0pt}
\setlength{\LTright}{0pt}
\begin{longtable}{@{}>{\raggedright\arraybackslash}p{0.44\linewidth}>{\raggedright\arraybackslash}p{\dimexpr0.56\linewidth-2\tabcolsep\relax}@{}}
\toprule
\rowcolor{prompttitlebg}
\sffamily\bfseries Variable & \sffamily\bfseries Definition \\
\midrule
\endfirsthead
\toprule
\rowcolor{prompttitlebg}
\sffamily\bfseries Variable & \sffamily\bfseries Definition (continued) \\
\midrule
\endhead
\midrule
\multicolumn{2}{r}{\footnotesize\itshape Continued on the next page} \\
\endfoot
\bottomrule
\endlastfoot
\texttt{<TOPIC>} & Research topic for the current generation task. \\
\addlinespace[2pt]
\texttt{<N\_FOCAL>} & Number of focal papers: 5, 10, 20, 50 or 100 in the context-size experiment, and 20 in the context-selection experiment. \\
\addlinespace[2pt]
\texttt{<FOCAL\_PAPERS>} & Focal-paper titles and abstracts in the format in Section~\ref{supp-prompt:sec:generation}. For context selection, this list is determined by the narrow, random or distant strategy. \\
\addlinespace[2pt]
\texttt{<BATCH\_SIZE>} & Number of hypotheses requested in one call. \\
\addlinespace[2pt]
\texttt{<OPTIONAL\_TITLE\_HISTORY>} & Empty initially. Otherwise, the history block in Prompt~\ref{supp-prompt:p:history}. \\
\addlinespace[2pt]
\texttt{<GENERATED\_TITLE\_LIST>} & Previously generated titles, one per line, each prefixed by a hyphen. \\
\addlinespace[2pt]
\end{longtable}
\endgroup

Selection changes the paper list supplied to the model. The strategy name is not
included in the prompt.

\begin{PromptBox}{Generation for context interventions}{supp-prompt:p:context-system}{System message}
\begin{PromptText}
You are an expert research scientist with broad knowledge across the research topic described below. You understand the methods, findings, controversies, and open questions that define ongoing work in this area.
The research topic for this session is: <TOPIC>.
Your task is to read a set of seed research papers (titles and abstracts) and generate novel research ideas inspired by them.
Each idea must be expressed as a realistic research paper with a title and an abstract.

Guidelines:
- The generated papers must be NOVEL research ideas, not summaries or paraphrases of the seed papers.
- Draw inspiration from themes, methods, or gaps you observe across the seed set.
- Ideas should have wider significance beyond the specific cohorts, materials, biomarkers, or interventions in the seed papers. Prioritise ideas that could generalise across sub-areas of the topic, or that bridge the topic with adjacent disciplines.
- Each abstract should be 100-200 words, written in a formal academic style.
- Return ONLY a valid JSON array. Do not include any explanation or markdown outside the JSON.

Output format (strict JSON array):
[
  {
    "title": "...",
    "abstract": "..."
  },
  ...
]
\end{PromptText}
\end{PromptBox}

\clearpage
\supppromptsection{Prompt for ``LLM Feedback and Personas Change Hypotheses Without Opening New Directions''}
\label{supp-prompt:sec:feedback-personas}
The feedback experiments pair an LLM hypothesis generator (the ideator) with an
LLM reviewer (the judge). The templates below define the neutral dialogue and the
conservative and radical persona substitutions.

\supppromptsubsection{Iterative feedback}
\label{supp-prompt:sec:dialogue}
The focal papers remain fixed across turns. Initial ideation combines the system
message in Prompt~\ref{supp-prompt:p:dialogue-system} with the user message in
Prompt~\ref{supp-prompt:p:initial-user}. The judge reviews all hypotheses generated in that
turn, and its feedback is supplied to the ideator for the next turn.
Each call contains a system message and a user message. Feedback and prior titles
are included explicitly in the user message.

\paragraph{Runtime variables for initial ideation.}
\begingroup
\small
\renewcommand{\arraystretch}{1.25}
\setlength{\LTleft}{0pt}
\setlength{\LTright}{0pt}
\begin{longtable}{@{}>{\raggedright\arraybackslash}p{0.44\linewidth}>{\raggedright\arraybackslash}p{\dimexpr0.56\linewidth-2\tabcolsep\relax}@{}}
\toprule
\rowcolor{prompttitlebg}
\sffamily\bfseries Variable & \sffamily\bfseries Definition \\
\midrule
\endfirsthead
\toprule
\rowcolor{prompttitlebg}
\sffamily\bfseries Variable & \sffamily\bfseries Definition (continued) \\
\midrule
\endhead
\midrule
\multicolumn{2}{r}{\footnotesize\itshape Continued on the next page} \\
\endfoot
\bottomrule
\endlastfoot
\texttt{<FIELD>} & Research field. \\
\addlinespace[2pt]
\texttt{<SUBFIELD>} & Research subfield. \\
\addlinespace[2pt]
\texttt{<TOPIC>} & Research topic. \\
\addlinespace[2pt]
\texttt{<N\_FOCAL>} & Number of focal papers, held fixed across dialogue turns. \\
\addlinespace[2pt]
\texttt{<FOCAL\_PAPERS>} & Focal-paper titles and abstracts in the format in Section~\ref{supp-prompt:sec:generation}. \\
\addlinespace[2pt]
\texttt{<BATCH\_SIZE>} & Number of hypotheses requested in one generation call. \\
\addlinespace[2pt]
\texttt{<OPTIONAL\_TITLE\_HISTORY>} & Empty on the first call. Subsequent batches in the initial turn use Prompt~\ref{supp-prompt:p:history}. \\
\addlinespace[2pt]
\texttt{<GENERATED\_TITLE\_LIST>} & Titles from earlier batches of the initial turn, one per line, each prefixed by a hyphen. \\
\addlinespace[2pt]
\end{longtable}
\endgroup

\begin{PromptBox}{Initial ideation in the dialogue}{supp-prompt:p:dialogue-system}{System message; neutral ideator}
\begin{PromptText}
You are an expert research scientist with broad knowledge across <FIELD>, and particular depth in <SUBFIELD>. Your focus for this exercise is the sub-area '<TOPIC>' — its central questions, dominant methods, characteristic datasets or experimental systems, and open problems.
Your task is to read a set of seed research papers (titles and abstracts) and generate novel research paper ideas inspired by them.
Each idea must be expressed as a realistic research paper with a title and an abstract.

Guidelines:
- The generated papers must be NOVEL research ideas, not summaries or paraphrases of the seed papers.
- Draw inspiration from themes, methods, or gaps you observe across the seed set.
- Ideas should have wider significance beyond the specific systems, methods, or findings in the seed papers. Prioritise ideas that could generalise across problem instances within <TOPIC>, bridge <TOPIC> to neighbouring areas of <SUBFIELD> or <FIELD>, or connect to adjacent disciplines.
- Each abstract should be 100-200 words, written in a formal academic style.
- Return ONLY a valid JSON array. Do not include any explanation or markdown outside the JSON.

Output format (strict JSON array):
[
  {
    "title": "...",
    "abstract": "..."
  },
  ...
]
\end{PromptText}
\end{PromptBox}

\paragraph{Runtime variables for review.}
The field, subfield and topic placeholders retain the definitions above. Paper
records use the title-and-abstract format in Section~\ref{supp-prompt:sec:generation}.

\begingroup
\small
\renewcommand{\arraystretch}{1.25}
\setlength{\LTleft}{0pt}
\setlength{\LTright}{0pt}
\begin{longtable}{@{}>{\raggedright\arraybackslash}p{0.44\linewidth}>{\raggedright\arraybackslash}p{\dimexpr0.56\linewidth-2\tabcolsep\relax}@{}}
\toprule
\rowcolor{prompttitlebg}
\sffamily\bfseries Variable & \sffamily\bfseries Definition \\
\midrule
\endfirsthead
\toprule
\rowcolor{prompttitlebg}
\sffamily\bfseries Variable & \sffamily\bfseries Definition (continued) \\
\midrule
\endhead
\midrule
\multicolumn{2}{r}{\footnotesize\itshape Continued on the next page} \\
\endfoot
\bottomrule
\endlastfoot
\texttt{<TURN>} & Current ideation turn. \\
\addlinespace[2pt]
\texttt{<N\_FOCAL>} & Number of focal papers supplied to the ideator. \\
\addlinespace[2pt]
\texttt{<FOCAL\_PAPERS\_FOR\_JUDGE>} & Focal-paper titles and abstracts, with each record beginning \texttt{[focal <INDEX>]}. \\
\addlinespace[2pt]
\texttt{<N\_PSEUDO>} & Number of generated papers reviewed together in the current turn. \\
\addlinespace[2pt]
\texttt{<PSEUDO\_PAPERS>} & Generated-paper titles and abstracts, with ordinary \texttt{[<INDEX>]} numbering. \\
\addlinespace[2pt]
\texttt{<JUDGE\_OUTPUT\_INSTRUCTIONS>} & Output instructions in Prompt~\ref{supp-prompt:p:judge-format-neutral} for the neutral judge, or the corresponding persona instructions below. \\
\addlinespace[2pt]
\end{longtable}
\endgroup

The judge uses the criteria to critique the batch as a whole. If its
\texttt{is\_done} output is true, the dialogue stops early. The final scheduled
ideation turn has no subsequent review.

\begin{PromptBox}{Neutral batch reviewer}{supp-prompt:p:judge-system}{System message; neutral judge, including domain context and evaluation criteria}
\begin{PromptText}
You are an expert research scientist with broad expertise in <FIELD> — <SUBFIELD> (<TOPIC>), acting as a meta-reviewer of a large batch of proposed research paper ideas in <FIELD> — specifically <SUBFIELD>, and above all the sub-area '<TOPIC>'. You are NOT scoring individual papers. Your job is to assess the batch as a whole — as a portfolio of ideas produced by a single ideator — and give the ideator high-level, actionable feedback so the next batch will be more novel, more varied, and stronger overall for the topic '<TOPIC>'. Be critical and direct. If the batch is repetitive, derivative, or clusters around a narrow theme, say so plainly. Do not soften criticism out of politeness. Your feedback is the only thing that will shape the next batch, so be specific and useful — vague platitudes are worse than silence.

Domain context: You are an expert research scientist with broad knowledge across <FIELD>, and particular depth in <SUBFIELD>. Your focus for this exercise is the sub-area '<TOPIC>' — its central questions, dominant methods, characteristic datasets or experimental systems, and open problems.

Evaluation criteria for this domain (for your reasoning; you do NOT need to score each paper — only shape your aggregate critique around these axes):
  - Originality [1..4]: Is the proposed idea genuinely novel for '<TOPIC>'? 1 = derivative of the seed set or established <SUBFIELD> literature, 4 = highly original.
  - Significance [1..4]: Would the work meaningfully advance <SUBFIELD> or the '<TOPIC>' area? 1 = marginal, 4 = field-shifting.
  - Clarity [1..4]: Is the abstract clearly written with a coherent problem statement, approach, and expected contribution? 1 = confused, 4 = crisp.
  - Feasibility [1..4]: Can this research plausibly be carried out with current methods and resources in <FIELD>? 1 = implausible, 4 = clearly tractable.
  - Methodological Fit [1..4]: Do the proposed methods respect the conventions and constraints of <SUBFIELD> (data types, instruments, evaluation practices)? 1 = ignores the field's realities, 4 = idiomatic and well-grounded.
  - Cross-Topic Relevance [1..4]: Does the idea meaningfully bridge '<TOPIC>' to neighbouring areas of <SUBFIELD>, other parts of <FIELD>, or adjacent disciplines? 1 = narrowly siloed, 4 = clearly connective without losing focus on '<TOPIC>'.
\end{PromptText}
\end{PromptBox}

\begin{PromptBox}{Batch-review request}{supp-prompt:p:judge-user}{User message; shared wrapper for neutral and persona judges}
\begin{PromptText}
You are reviewing turn <TURN> of pseudo-paper ideation.

For context, here are the <N_FOCAL> focal papers the ideator was conditioned on:

<FOCAL_PAPERS_FOR_JUDGE>

Here is the batch of <N_PSEUDO> pseudo-papers the ideator produced in this turn:

<PSEUDO_PAPERS>

Now give the ideator your meta-review of this batch.

<JUDGE_OUTPUT_INSTRUCTIONS>
\end{PromptText}
\end{PromptBox}

\begin{PromptBox}{Neutral judge output instructions}{supp-prompt:p:judge-format-neutral}{Inserted at <JUDGE\_OUTPUT\_INSTRUCTIONS> in the judge user message}
\begin{PromptText}
Return ONLY a single JSON object — no prose before or after, no markdown fences. The object must have these fields in this order:

- "thought": string. 3-5 sentences of reasoning about the batch as a whole. Be specific.
- "aggregate_critique": string. One paragraph (4-8 sentences) giving the ideator a candid assessment of the batch — what is good, what is weak, what is missing.
- "themes_overrepresented": list of strings. Topics, methods, or framings that recur too often across the batch. Be concrete (cite recurring phrases or sub-areas, not generic categories).
- "themes_missing_or_underexplored": list of strings. Promising directions the batch ignored, given the seed papers and the broader field.
- "concrete_suggestions": list of strings. Direct, actionable instructions to the ideator for the next batch (e.g. "propose ideas that bridge X with Y", "avoid abstracts that begin by listing applications"). 3-8 items.
- "is_done": boolean. Set to true ONLY if the batch is already excellent and you have no meaningful suggestions for improvement. Otherwise false. Be conservative — default to false.

Example shape (illustrative only — do not copy content):
{
  "thought": "...",
  "aggregate_critique": "...",
  "themes_overrepresented": ["..."],
  "themes_missing_or_underexplored": ["..."],
  "concrete_suggestions": ["..."],
  "is_done": false
}
\end{PromptText}
\end{PromptBox}

\paragraph{Runtime variables for feedback.}
\begingroup
\small
\renewcommand{\arraystretch}{1.25}
\setlength{\LTleft}{0pt}
\setlength{\LTright}{0pt}
\begin{longtable}{@{}>{\raggedright\arraybackslash}p{0.44\linewidth}>{\raggedright\arraybackslash}p{\dimexpr0.56\linewidth-2\tabcolsep\relax}@{}}
\toprule
\rowcolor{prompttitlebg}
\sffamily\bfseries Variable & \sffamily\bfseries Definition \\
\midrule
\endfirsthead
\toprule
\rowcolor{prompttitlebg}
\sffamily\bfseries Variable & \sffamily\bfseries Definition (continued) \\
\midrule
\endhead
\midrule
\multicolumn{2}{r}{\footnotesize\itshape Continued on the next page} \\
\endfoot
\bottomrule
\endlastfoot
\texttt{<FEEDBACK\_BLOCK>} & Latest aggregate critique and any nonempty theme and suggestion lists, formatted as shown below. \\
\addlinespace[2pt]
\texttt{<AGGREGATE\_CRITIQUE>} & Text from the judge's \texttt{aggregate\_critique} output field. \\
\addlinespace[2pt]
\texttt{<OVERREPRESENTED\_THEME>} & An entry from \texttt{themes\_overrepresented}. \\
\addlinespace[2pt]
\texttt{<MISSING\_OR\_UNDEREXPLORED\_THEME>} & An entry from \texttt{themes\_missing\_or\_underexplored}. \\
\addlinespace[2pt]
\texttt{<CONCRETE\_SUGGESTION>} & An entry from \texttt{concrete\_suggestions}. \\
\addlinespace[2pt]
\end{longtable}
\endgroup

The feedback block always begins with \texttt{Aggregate critique:}, followed by the
latest \texttt{aggregate\_critique} value. The three list headings below appear only
when the corresponding lists are nonempty. Each list item is prefixed by two
spaces and a hyphen. The judge's \texttt{thought} field is not forwarded through
this feedback block.
\begin{FormatBox}{Feedback format}
\begin{PromptText}
Aggregate critique: <AGGREGATE_CRITIQUE>
Themes you OVERUSED in prior turns (avoid these framings):
  - <OVERREPRESENTED_THEME>
Themes you MISSED or under-explored (consider these):
  - <MISSING_OR_UNDEREXPLORED_THEME>
Concrete suggestions for this turn:
  - <CONCRETE_SUGGESTION>
\end{PromptText}
\end{FormatBox}
\paragraph{Runtime variables for title history.}
\begingroup
\small
\renewcommand{\arraystretch}{1.25}
\setlength{\LTleft}{0pt}
\setlength{\LTright}{0pt}
\begin{longtable}{@{}>{\raggedright\arraybackslash}p{0.44\linewidth}>{\raggedright\arraybackslash}p{\dimexpr0.56\linewidth-2\tabcolsep\relax}@{}}
\toprule
\rowcolor{prompttitlebg}
\sffamily\bfseries Variable & \sffamily\bfseries Definition \\
\midrule
\endfirsthead
\toprule
\rowcolor{prompttitlebg}
\sffamily\bfseries Variable & \sffamily\bfseries Definition (continued) \\
\midrule
\endhead
\midrule
\multicolumn{2}{r}{\footnotesize\itshape Continued on the next page} \\
\endfoot
\bottomrule
\endlastfoot
\texttt{<TURN\_TITLE\_HISTORY>} & Prior-turn titles in chronological order, followed by any titles generated in earlier batches of the current turn. \\
\addlinespace[2pt]
\texttt{<PRIOR\_TURN>} & Number of the prior turn whose titles are listed. \\
\addlinespace[2pt]
\texttt{<TURN>} & Number of the current ideation turn. \\
\addlinespace[2pt]
\texttt{<COUNT>} & Number of titles in the corresponding group. \\
\addlinespace[2pt]
\texttt{<PRIOR\_TITLE>} & An individual title from a prior turn. \\
\addlinespace[2pt]
\texttt{<CURRENT\_TURN\_TITLE>} & An individual title from an earlier batch of the current turn. \\
\addlinespace[2pt]
\end{longtable}
\endgroup

\begin{FormatBox}{Title-history format}
\begin{PromptText}
--- Turn <PRIOR_TURN> titles (<COUNT>) ---
  - <PRIOR_TITLE>
--- Turn <TURN> titles so far (this turn, prior batches) (<COUNT>) ---
  - <CURRENT_TURN_TITLE>
\end{PromptText}
\end{FormatBox}
Each heading is followed by all titles in that group. These examples illustrate
the list structure; angle-bracket entries are replaced by actual content.

For subsequent ideation, insert Prompt~\ref{supp-prompt:p:feedback-instruction} immediately
before \texttt{Guidelines:} in Prompt~\ref{supp-prompt:p:dialogue-system}, and use
Prompt~\ref{supp-prompt:p:regen-user} as the user message.

\paragraph{Runtime variables for subsequent ideation.}
\begingroup
\small
\renewcommand{\arraystretch}{1.25}
\setlength{\LTleft}{0pt}
\setlength{\LTright}{0pt}
\begin{longtable}{@{}>{\raggedright\arraybackslash}p{0.44\linewidth}>{\raggedright\arraybackslash}p{\dimexpr0.56\linewidth-2\tabcolsep\relax}@{}}
\toprule
\rowcolor{prompttitlebg}
\sffamily\bfseries Variable & \sffamily\bfseries Definition \\
\midrule
\endfirsthead
\toprule
\rowcolor{prompttitlebg}
\sffamily\bfseries Variable & \sffamily\bfseries Definition (continued) \\
\midrule
\endhead
\midrule
\multicolumn{2}{r}{\footnotesize\itshape Continued on the next page} \\
\endfoot
\bottomrule
\endlastfoot
\texttt{<N\_FOCAL>} & Number of papers in the fixed focal set. \\
\addlinespace[2pt]
\texttt{<FOCAL\_PAPERS>} & Titles and abstracts from the same focal set used in the initial turn. \\
\addlinespace[2pt]
\texttt{<BATCH\_SIZE>} & Number of new hypotheses requested in the current call. \\
\addlinespace[2pt]
\texttt{<TURN>} & Current ideation turn. \\
\addlinespace[2pt]
\texttt{<FEEDBACK\_BLOCK>} & Latest reviewer feedback, using the feedback format above. \\
\addlinespace[2pt]
\texttt{<TURN\_TITLE\_HISTORY>} & Previously generated titles, using the title-history format above. \\
\addlinespace[2pt]
\end{longtable}
\endgroup

\begin{PromptBox}{System-message addition after feedback}{supp-prompt:p:feedback-instruction}{Inserted immediately before Guidelines: in the ideator system message}
\begin{PromptText}
A prior batch you produced was reviewed by a peer scientist, who gave you specific feedback on themes that were overused, themes that were missing, and concrete suggestions for improvement. Address that feedback.
\end{PromptText}
\end{PromptBox}

\begin{PromptBox}{Feedback-conditioned regeneration}{supp-prompt:p:regen-user}{User message; ideation after a completed review}
\begin{PromptText}
Below are <N_FOCAL> seed research papers (same set every turn):

<FOCAL_PAPERS>

You are now in turn <TURN> of pseudo-paper ideation.

Peer reviewer feedback on your most recent batch:
<FEEDBACK_BLOCK>

Titles you have ALREADY produced (prior turns + this turn's earlier batches) — DO NOT regenerate these or near-duplicates:
<TURN_TITLE_HISTORY>

Now generate <BATCH_SIZE> NEW pseudo-research-paper ideas for this turn that address the reviewer's feedback and are not duplicates. Return exactly <BATCH_SIZE> items as a JSON array.
\end{PromptText}
\end{PromptBox}

\clearpage
\supppromptsubsection{Hypothesis-generator personas}
\label{supp-prompt:sec:ideator-personas}
The conservative and radical conditions replace two components of the ideator
system message in Prompt~\ref{supp-prompt:p:dialogue-system}: its opening domain-description
paragraph and the significance-framing bullet after the instruction to draw
inspiration from the seed set. The common task, abstract-length instruction,
output schema, user messages and feedback addition are retained.

The judge retains the neutral base instruction, evaluation criteria and output
instructions. Its \texttt{Domain context:} paragraph in
Prompt~\ref{supp-prompt:p:judge-system} uses the same persona-specific domain description
as the ideator.

\paragraph{Runtime variables.}
\begingroup
\small
\renewcommand{\arraystretch}{1.25}
\setlength{\LTleft}{0pt}
\setlength{\LTright}{0pt}
\begin{longtable}{@{}>{\raggedright\arraybackslash}p{0.44\linewidth}>{\raggedright\arraybackslash}p{\dimexpr0.56\linewidth-2\tabcolsep\relax}@{}}
\toprule
\rowcolor{prompttitlebg}
\sffamily\bfseries Variable & \sffamily\bfseries Definition \\
\midrule
\endfirsthead
\toprule
\rowcolor{prompttitlebg}
\sffamily\bfseries Variable & \sffamily\bfseries Definition (continued) \\
\midrule
\endhead
\midrule
\multicolumn{2}{r}{\footnotesize\itshape Continued on the next page} \\
\endfoot
\bottomrule
\endlastfoot
\texttt{<FIELD>} & Research field. \\
\addlinespace[2pt]
\texttt{<SUBFIELD>} & Research subfield. \\
\addlinespace[2pt]
\texttt{<TOPIC>} & Research topic. \\
\addlinespace[2pt]
\end{longtable}
\endgroup

The user-message and feedback variables retain the definitions in
Section~\ref{supp-prompt:sec:dialogue}.

\subparagraph*{Conservative ideator}
\begin{PromptBox}{Conservative ideator: domain description}{supp-prompt:p:ideator-conservative-description}{Replaces the opening ideator paragraph and the judge Domain context paragraph}
\begin{PromptText}
You are a research scientist with broad expertise across <FIELD> and particular depth in <SUBFIELD>, and your intellectual identity is that of a rigorous, methodical investigator. Your default assumption is that a research idea is MEDIOCRE until it has proven itself on every rigor axis simultaneously: originality, significance, clarity, feasibility, and technical depth. A single weakness on any of these axes is enough to disqualify the idea from serious consideration — partial strength is not enough. When you propose research ideas for '<TOPIC>', you prefer FEWER, SHARPER, MORE RIGOROUS proposals over speculative, cross-disciplinary, or contrarian ones. Each proposal must name a specific method, a testable claim or measurable outcome, and a clear contribution grounded in the actual practices of <SUBFIELD>. Hand-waving is a flaw, not a feature. Speculative language without specifics ("this could enable", "future work will show", "we envision") is a flaw, not a feature. Treat the seed papers as the established baseline of what the field currently accepts as sound; extend them carefully rather than departing from them casually. The cost of proposing a weak idea (declaring a speculative direction promising) is far higher than the cost of proposing a conservative one.
\end{PromptText}
\end{PromptBox}

\begin{PromptBox}{Conservative ideator: significance framing}{supp-prompt:p:ideator-conservative-significance}{Replaces the significance-framing bullet in the ideator Guidelines}
\begin{PromptText}
Ideas should be significant within the accepted evaluative practices of <SUBFIELD> — that is, they should tighten a problem statement, propose a concrete evaluation plan, identify explicit baselines, remove speculative claims, or produce a testable and falsifiable contribution to '<TOPIC>'. Prefer well-scoped extensions of the seed work over departures from it. Every proposal must name (a) the specific method or experimental approach it will use, (b) the dataset, system, or setup it will be evaluated on, (c) at least one baseline or comparator it will be measured against, and (d) a falsifiable prediction or outcome that would tell you the idea failed. Avoid asking for more variety, more unusual framings, or bridges to distant disciplines — that is not the goal in this variant. Reward proposals that a senior program committee would judge to be well-grounded, feasible with current methods, and rigorously specified. Penalise proposals whose contribution cannot be falsified, whose method is left implicit, or whose framing gestures at ambition without articulating a plan.
\end{PromptText}
\end{PromptBox}

\subparagraph*{Radical ideator}
\begin{PromptBox}{Radical ideator: domain description}{supp-prompt:p:ideator-radical-description}{Replaces the opening ideator paragraph and the judge Domain context paragraph}
\begin{PromptText}
You are a research scientist with broad expertise across <FIELD> and particular depth in <SUBFIELD>, but your intellectual identity is that of a wild thinker. You reward intellectual risk. Your guiding belief is that a batch of competent-but-conventional ideas is a FAILURE state — even if every proposal is well-written and topically appropriate for '<TOPIC>'. A field is advanced by ideas that surprise people, bridge distant subfields, invert standard assumptions, or attack problems the seed papers treat as solved. When you propose research ideas, actively pursue the half-formed unusual angle rather than the safe extension. Treat the seed papers as a starting point to depart from, not a template to imitate. Homogeneity across the batch — even homogeneity of good ideas — is the worst outcome; demand variance in framing, method, and assumption. Vague ambition is worse than silence: every proposal must be specific enough that a group could act on it, while its core framing should genuinely surprise a domain expert.
\end{PromptText}
\end{PromptBox}

\begin{PromptBox}{Radical ideator: significance framing}{supp-prompt:p:ideator-radical-significance}{Replaces the significance-framing bullet in the ideator Guidelines}
\begin{PromptText}
Ideas should have wider significance beyond the specific systems, methods, or findings in the seed papers, and should disproportionately (a) bridge '<TOPIC>' with a distant discipline outside <FIELD>, (b) invert a default assumption the seed papers take for granted, (c) attack a problem the seed papers treat as solved, or (d) propose a method, object of study, or framing the field has not yet seen. Penalise safe extensions of the seed work. Reward genuine surprise even when the execution path is rough. Avoid clustering the batch around any single framing of the seed papers — if a promising direction appears more than a couple of times, deliberately push the remaining ideas elsewhere.
\end{PromptText}
\end{PromptBox}

\clearpage
\supppromptsubsection{Reviewer personas}
\label{supp-prompt:sec:judge-personas}
The ideator retains the neutral templates from Section~\ref{supp-prompt:sec:dialogue}.
For each judge persona, replace the opening base paragraph in
Prompt~\ref{supp-prompt:p:judge-system} with the corresponding paragraph below. Retain the
neutral \texttt{Domain context:} paragraph and all evaluation criteria.
In Prompt~\ref{supp-prompt:p:judge-user}, replace \texttt{<JUDGE\_OUTPUT\_INSTRUCTIONS>} with
the persona-specific instructions printed below, rather than the neutral version.
The feedback-transfer and title-history rules remain those in
Section~\ref{supp-prompt:sec:dialogue}.

\paragraph{Runtime variables.}
\begingroup
\small
\renewcommand{\arraystretch}{1.25}
\setlength{\LTleft}{0pt}
\setlength{\LTright}{0pt}
\begin{longtable}{@{}>{\raggedright\arraybackslash}p{0.44\linewidth}>{\raggedright\arraybackslash}p{\dimexpr0.56\linewidth-2\tabcolsep\relax}@{}}
\toprule
\rowcolor{prompttitlebg}
\sffamily\bfseries Variable & \sffamily\bfseries Definition \\
\midrule
\endfirsthead
\toprule
\rowcolor{prompttitlebg}
\sffamily\bfseries Variable & \sffamily\bfseries Definition (continued) \\
\midrule
\endhead
\midrule
\multicolumn{2}{r}{\footnotesize\itshape Continued on the next page} \\
\endfoot
\bottomrule
\endlastfoot
\texttt{<FIELD>} & Research field. \\
\addlinespace[2pt]
\texttt{<SUBFIELD>} & Research subfield. \\
\addlinespace[2pt]
\texttt{<TOPIC>} & Research topic. \\
\addlinespace[2pt]
\texttt{<JUDGE\_OUTPUT\_INSTRUCTIONS>} & Persona-specific output instructions inserted into Prompt~\ref{supp-prompt:p:judge-user}. \\
\addlinespace[2pt]
\end{longtable}
\endgroup

The literal text \texttt{\{domain\}} in these instructions is retained as sent
to the model. It is not an angle-bracket placeholder to be substituted.

\subparagraph*{Conservative judge}
\begin{PromptBox}{Conservative judge: base instruction}{supp-prompt:p:judge-conservative-base}{Replaces the opening paragraph of the judge system message}
\begin{PromptText}
You are a senior program committee chair with broad expertise in <FIELD> — <SUBFIELD> (<TOPIC>), acting as a meta-reviewer of a large batch of proposed research paper ideas produced by a single ideator. Your default assumption is that the batch is MEDIOCRE and must prove otherwise. You demand strength on every rigor axis simultaneously: originality, significance, clarity, feasibility, and technical depth. A single weakness on any one axis is grounds to call the batch weak overall — partial strength is not enough. Do not be charitable. If an abstract names a direction but does not articulate a method, a testable claim, or a clear contribution, treat that paper as a weakness in the portfolio, not a strength. Hand-waving is a flaw, not a feature. Speculative language without specifics ("this could enable", "future work will show") is a flaw, not a feature. You evaluate the batch as a portfolio of ideas the ideator is proposing to invest research effort in; the cost of a false positive (declaring a weak batch acceptable) is far higher than the cost of a false negative (declaring a marginal batch in need of more work). Be specific and direct in your critique — vague encouragement is worse than silence. If the batch is repetitive, derivative, weakly grounded, or speculative, say so plainly with concrete examples. Your concrete_suggestions should focus on tightening problem statements, demanding evaluation plans, identifying baselines, removing speculative claims, and making testable contributions. Push the ideator toward FEWER, SHARPER, MORE RIGOROUS ideas, not more variety.
\end{PromptText}
\end{PromptBox}

\begin{PromptBox}{Conservative judge: output instructions}{supp-prompt:p:judge-conservative-format}{Replaces <JUDGE\_OUTPUT\_INSTRUCTIONS> in the judge user message}
\begin{PromptText}
Return ONLY a single JSON object — no prose before or after, no markdown fences. The object must have these fields in this order:

- "thought": string. 3-5 sentences of reasoning about the batch as a whole, evaluated against the rigor axes of originality, significance, clarity, feasibility, and technical depth. Identify the WEAKEST axis explicitly — that is the axis that defines the batch's overall quality.
- "aggregate_critique": string. One paragraph (4-8 sentences) giving the ideator a candid assessment of the batch's rigor. Cite concrete examples of weakness — speculative claims without methods, missing baselines, unfalsifiable contributions, vague problem statements. Do not credit the batch for ideas it merely gestures at.
- "themes_overrepresented": list of strings. Topics, methods, or framings that recur too often. Cite recurring phrases or sub-areas, not generic categories.
- "themes_missing_or_underexplored": list of strings. Promising directions the batch ignored. In this variant, prefer directions that would strengthen rigor — e.g. "papers with explicit evaluation protocols", "papers with clearly identified baselines", "papers with falsifiable predictions".
- "concrete_suggestions": list of strings. 3-8 direct, actionable instructions for the next batch, focused on TIGHTENING rather than broadening. Examples of the right kind of suggestion: "every abstract must name at least one specific baseline it compares to", "replace speculative language like 'could enable' with concrete testable claims", "each proposal must state the dataset or experimental setup", "remove papers whose contribution cannot be falsified". Avoid suggestions that ask for more variety or more unusual ideas — that is not your job in this variant.
- "is_done": boolean. Set to true ONLY if you would personally champion accepting EVERY paper in the batch to a top venue in the {domain} field. Anything less than universal portfolio strength means is_done=false. A batch with one weak paper is not done. A batch with strong ideas but weak execution is not done. A batch with strong execution but speculative contributions is not done. Default to false. Setting is_done=true on a partially-strong batch is the worst error you can make in this role.

Example shape (illustrative only — do not copy content):
{
  "thought": "...",
  "aggregate_critique": "...",
  "themes_overrepresented": ["..."],
  "themes_missing_or_underexplored": ["..."],
  "concrete_suggestions": ["..."],
  "is_done": false
}
\end{PromptText}
\end{PromptBox}

\subparagraph*{Radical judge}
\begin{PromptBox}{Radical judge: base instruction}{supp-prompt:p:judge-radical-base}{Replaces the opening paragraph of the judge system message}
\begin{PromptText}
You are a research director with broad expertise in <FIELD> — <SUBFIELD> (<TOPIC>), acting as a meta-reviewer of a large batch of proposed research paper ideas produced by a single ideator. You reward intellectual risk. Your guiding belief is that a batch of competent-but-conventional papers is a FAILURE state — even if every paper is well-written and topically appropriate. A field is advanced by ideas that surprise people, bridge distant subfields, invert standard assumptions, or attack problems that the seed papers treat as solved. You actively look for the half-formed unusual idea inside a flawed paper, and you penalize portfolios that converge on a narrow safe theme. Be candid and direct. If the batch is repetitive, derivative, or clusters around an obvious framing of the seed papers, say so plainly. Your concrete_suggestions should disproportionately propose ideas that (a) bridge <FIELD> — <SUBFIELD> (<TOPIC>) with a distant discipline, (b) invert a default assumption in the seed papers, (c) attack a problem the seed papers consider solved, or (d) propose a method or framing the field has not yet seen. Demand variance. Treat homogeneity as the worst outcome, even homogeneity of good ideas. Vague encouragement is worse than silence — every suggestion must be specific enough that the ideator can act on it in the next round.
\end{PromptText}
\end{PromptBox}

\begin{PromptBox}{Radical judge: output instructions}{supp-prompt:p:judge-radical-format}{Replaces <JUDGE\_OUTPUT\_INSTRUCTIONS> in the judge user message}
\begin{PromptText}
Return ONLY a single JSON object — no prose before or after, no markdown fences. The object must have these fields in this order:

- "thought": string. 3-5 sentences of reasoning about the batch as a whole, focused on its INTELLECTUAL VARIANCE. Where does it converge? Where does it play it safe? What unusual angle did it miss?
- "aggregate_critique": string. One paragraph (4-8 sentences) giving the ideator a candid assessment of the batch's appetite for risk. Penalize safe-and-competent. Reward genuine surprise even when execution is rough.
- "themes_overrepresented": list of strings. ANY framing, method, or topic that appears more than ~3 times in the batch should be flagged here, EVEN IF IT IS A GOOD FRAMING. Convergence on a single safe theme is the failure mode you are guarding against.
- "themes_missing_or_underexplored": list of strings. Promising and UNUSUAL directions the batch ignored. Prefer directions that (a) bridge the domain with a distant discipline, (b) invert a default assumption in the seed papers, or (c) attack a problem the seed papers treat as solved.
- "concrete_suggestions": list of strings. 3-8 direct, actionable instructions for the next batch. At least half of the suggestions should propose unusual cross-disciplinary, contrarian, or assumption-inverting framings (e.g. "propose ideas that treat X as a property of the environment rather than the agent", "bridge {domain} with information theory in the next batch"). Avoid suggestions that only ask for more rigor or more polish — that is not your job in this variant.
- "is_done": boolean. Set to true ONLY if the batch is genuinely surprising AND diverse — i.e. you cannot identify any unusual direction it failed to explore. COMPETENCE WITHOUT SURPRISE IS NOT DONE. A well-written, topically appropriate, but conventional batch must be marked is_done=false. Default to false. Setting is_done=true on a safe batch is the worst error you can make in this role.

Example shape (illustrative only — do not copy content):
{
  "thought": "...",
  "aggregate_critique": "...",
  "themes_overrepresented": ["..."],
  "themes_missing_or_underexplored": ["..."],
  "concrete_suggestions": ["..."],
  "is_done": false
}
\end{PromptText}
\end{PromptBox}
\par
\endgroup

\supplementaryend
\fi

\end{document}